\documentclass{article}
\usepackage{iclr2025_conference,times}
\usepackage[protrusion=false,expansion=false]{microtype}
\usepackage{etoolbox}
\SetTracking{encoding=OT1,family=ptm}{-25}

\usepackage{amsmath,amsfonts,bm}

\def\eqref#1{equation~\ref{#1}}

\def\1{\bm{1}}

\DeclareMathAlphabet{\mathsfit}{\encodingdefault}{\sfdefault}{m}{sl}
\SetMathAlphabet{\mathsfit}{bold}{\encodingdefault}{\sfdefault}{bx}{n}

\usepackage{hyperref}
\hypersetup{hidelinks}
\usepackage[all]{hypcap}
\usepackage{url}
\usepackage{graphicx}
\usepackage{float}
\usepackage{wrapfig}
\usepackage{placeins}
\usepackage{array}
\usepackage{multirow}
\usepackage{makecell}
\usepackage{colortbl}
\usepackage{booktabs}
\usepackage[labelsep=period]{caption}
\bibpunct{(}{)}{,}{a}{,}{,}
\newcolumntype{L}[1]{>{\raggedright\arraybackslash}p{#1}}
\newcolumntype{C}[1]{>{\centering\arraybackslash}p{#1}}
\definecolor{otInk}{HTML}{18242C}
\definecolor{otSoft}{HTML}{F4F6F6}
\definecolor{otGold}{HTML}{F3B63F}
\definecolor{otBlue}{HTML}{3D86AF}
\definecolor{otCoral}{HTML}{F26749}
\definecolor{otTeal}{HTML}{63B8B0}
\definecolor{otPurple}{HTML}{8E72B5}
\definecolor{otGreen}{HTML}{0DA56B}

\title{OpenMAS-GCom. A Diagnostic Benchmark for Graph-enhanced Multi-Agent Systems}

\iclrfinalcopy
\author{Kairui Yang\textsuperscript{*}, Xunkai Li\textsuperscript{*}, Kaixiang Zhang,\\
\bfseries Minghao An, Zekai Chen, Yuxuan Ba, Rong-Hua Li\\
\normalfont\small\textsuperscript{*}Equal contribution.}
\hypersetup{pdfauthor={Kairui Yang, Xunkai Li, Kaixiang Zhang, Minghao An, Zekai Chen, Yuxuan Ba, Rong-Hua Li},pdftitle={OpenMAS-GCom. A Diagnostic Benchmark for Graph-enhanced Multi-Agent Systems}}

\begin{document}
\maketitle
\fancyhead{}
\renewcommand{\headrulewidth}{0pt}
\raggedbottom

\begin{abstract}
Graph-enhanced multi-agent systems (G-MAS) coordinate large language model agents through communication graphs and role assignments, which determine how agents exchange information and divide responsibilities.
However, final-score comparisons across systems combine differences in models, communication patterns, roles, and computation costs, making performance differences difficult to attribute to specific communication structures, role assignments, and information flows.
To address this evaluation attribution problem, we introduce OpenMAS-GCom, a benchmark for diagnosing how these components affect G-MAS performance through controlled interventions.
We represent systems through collaboration units, communication links, shared intermediate information, and execution rules.
OpenMAS-GCom compares original systems with versions modified by changing one component while keeping tasks, models, prompts, and budget limits fixed.
We rewire communication edges, remove specialist or critic agents, replace intermediate messages with incorrect content, and disable workers during execution.
The benchmark evaluates 17 single-agent, ordinary multi-agent, and graph-enhanced configurations on 29 datasets across six domains.
We add 400 G-MAS-Complex tasks requiring agents to combine information from multiple documents, resolve conflicting records, and return specified values with source identifiers.
Experiments show larger mean losses after specialist removal than after critic removal, different performance degradation under incorrect messages and worker failures despite similar original scores, and different configurations achieving the highest accuracy and accuracy per token on G-MAS-Complex.
\end{abstract}

\section{Introduction}
\label{sec:introduction}

Large language model (LLM) agents collaborate by assigning roles and exchanging intermediate results to solve mathematical reasoning, code generation, and question answering tasks~\citep{wu2023autogen,du2023multiagentdebate}.
In graph-enhanced multi-agent systems (G-MAS), communication graphs determine which agents exchange information, while roles determine which agents generate, verify, or integrate solutions~\citep{zhuge2024gptswarm,zhang2024gdesigner}.
For example, one agent proposes an answer, another checks it, and a third combines intermediate results.
Performance therefore depends on both the underlying language model and how agents communicate and divide responsibilities.

Existing evaluations mainly compare final task scores across systems. Recent studies analyze information propagation across communication topologies, while MultiAgentBench evaluates task completion under different multi-agent settings~\citep{shen2025eib,zhu2025multiagentbench}.
A higher score may accompany a stronger model, more agents, different communication links or roles, and greater computation costs.
These factors often change together across methods, making performance differences difficult to attribute to specific organizational choices.
This creates an evaluation attribution problem for G-MAS. How do communication structures, role composition, and information flow affect performance when other evaluation settings are held fixed?

To address this problem, we present OpenMAS-GCom, a diagnostic benchmark using controlled organizational interventions.
We represent each system through collaboration units, communication links, shared intermediate information, and execution rules.
We compare original and modified systems after changing one component, with task inputs, model settings, prompts, and computation budget limits fixed.
We evaluate 17 single-agent, ordinary multi-agent, and graph-enhanced configurations on 29 datasets across six domains through a common interface that records outputs, intermediate messages, executed units, and resource use.
We introduce 400 G-MAS-Complex tasks to examine how agents combine information from multiple documents, resolve conflicting records, and return specified values with source identifiers.
Four intervention protocols rewire communication edges while preserving node degrees, remove specialist or critic agents, replace intermediate messages with incorrect content, and disable workers during execution.

\textbf{Our Contributions.}
(1) \textbf{Comprehensive Benchmark.} We integrate datasets, baseline configurations, and four intervention protocols for evaluating changes to communication, roles, intermediate information, and worker availability under shared settings.
(2) \textbf{Valuable Insights.} Across five datasets, specialist removal reduces the mean score by 3.58 percentage points, compared with 0.58 for critic removal. MAD and VeriMap start with similar accuracy but respond differently to incorrect messages. On G-MAS-Complex, different configurations achieve the highest accuracy and accuracy per token.
(3) \textbf{Open-sourced Benchmark Library.} We organize task loading, method execution, scoring, and interventions into reusable modules. Run identifiers link configurations, predictions, intervention settings, and costs to support reproducible evaluation.

\setlength{\parskip}{3.5pt}
\section{Problem Statement}
\label{sec_problem_statement}

OpenMAS-GCom connects task evaluation with organizational diagnosis through two complementary pipelines. The collaboration pipeline produces predictions and execution records through a common interface. The diagnostic pipeline measures how these outcomes change when selected organizational components are modified.

\subsection{End-to-End Collaboration Pipeline}
\label{sec_end_to_end_collaboration}

Given a task domain $T$ with evaluation set $D_T=\{(x_i,y_i)\}_{i=1}^{N}$, a method $F$ specifies organization construction and execution. Under configuration $\lambda$, its construction procedure produces $\mathcal{O}_i=F_{\mathrm{build}}(x_i,\lambda)$. A fixed design uses the same organization across inputs, while task-dependent construction can change its units and connections. Each organization has the form
\begin{equation}
\mathcal{O}=(V,E,S,\pi),
\label{eq_organization}
\end{equation}
where $V$ specifies runnable collaboration units, $E$ specifies permitted directed communication, $S$ holds shared information and intermediate states, and $\pi$ governs activation, routing, state updates, and termination. A unit declares its role, local state, and input and output schemas. Agent units form a subset $V_A\subseteq V$, alongside tool or control units. These components separate participation, communication, information access, and execution decisions.

For input $x_i$, sequential workflows activate units in order, parallel workflows activate several eligible units, and debate repeats message exchange through the policy. Graph-based execution uses the configured relations to select permitted transfers and activations. The runner records the trace
\begin{equation}
\tau_i=((v_t,m_t,a_t))_{t=1}^{T_i},
\label{eq_trace}
\end{equation}
where $v_t$ is the executed unit, $m_t$ is its available message state, and $a_t$ is its output or control action. The final output $o_i$ is extracted from the completed trace. Associated records contain token usage, model calls, latency, executed units and relations, retries, and termination status.

The representation captures the organizational properties exposed to the runner. Method-specific prompts, learned parameters, tool definitions, and controller logic remain in the configuration and policy. Shared-state permissions specify the information each unit may read or write. The configured organization describes available units and relations, while the trace identifies those used for an input. This distinction connects the comparison interface with configuration-specific execution details.

The task score is
\begin{equation}
s(F,\mathcal{O}_T,T)=\frac{1}{N}\sum_{i=1}^{N}M_T(P(o_i),y_i),
\label{eq_task_score}
\end{equation}
where $P$ parses the output, $M_T$ is the task-specific metric, and $\mathcal{O}_T=\{\mathcal{O}_i\}_{i=1}^{N}$ collects the task-specific organizations. The parser and metric connect execution with comparable task scores.

\subsection{Two-Stage Diagnostic Pipeline}
\label{sec_two_stage_diagnosis}

The reference stage executes the original organization and records predictions, traces, and costs. The intervention stage modifies one organizational component and repeats the evaluation on the same tasks. Shared controls are
\begin{equation}
\mathcal{C}=\{D_T,\Theta,M_T,P\},
\label{eq_shared_control}
\end{equation}
where $\Theta$ contains the backend, prompts, decoding and routing settings, stopping rules, and task-level execution and communication budget limits retained within each pair. For intervention type $r$, the modified organization is
\begin{equation}
\mathcal{O}'_{r,i}=I_r(\mathcal{O}_i,\Delta\mathcal{O}_{r,i}).
\label{eq_intervention}
\end{equation}
The paired intervention response compares the resulting task scores
\begin{equation}
\Delta_r(F,T)=s(F,\mathcal{O}'_{r,T},T)-s(F,\mathcal{O}_T,T).
\label{eq_diagnostic_effect}
\end{equation}
Here $\mathcal{O}'_{r,T}=\{\mathcal{O}'_{r,i}\}_{i=1}^{N}$ collects the modified organizations. The paired score change measures the average response to intervention $r$ under the controls in Equation~\ref{eq_shared_control}. A negative value indicates lower performance after modification. Its unit of analysis is the complete modified execution, including subsequent message processing and actions. Structure, node, information, and execution interventions modify relations, units, intermediate messages, and worker availability, respectively.

\section{Benchmark Design}
\label{sec_benchmark_design}

\begin{figure}[t]
  \centering
  \includegraphics[width=\textwidth]{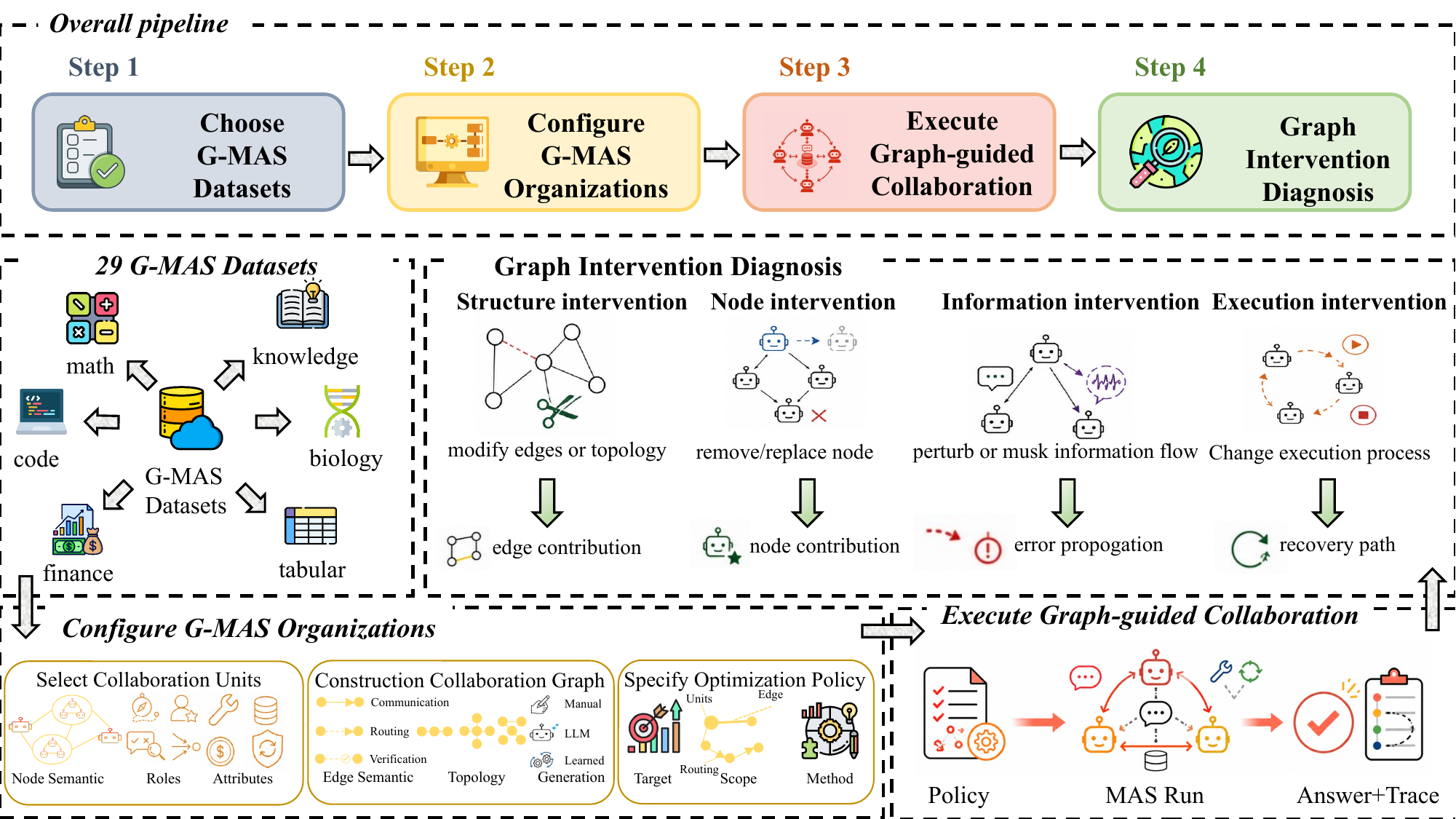}
  \caption{OpenMAS-GCom model framework. The benchmark selects datasets, configures G-MAS organizations, executes graph guided collaboration, and applies four graph intervention protocols.}
  \label{fig_opengmas_framework}
\end{figure}

OpenMAS-GCom integrates task selection, organization configuration, collaborative execution, and paired diagnosis. Figure~\ref{fig_opengmas_framework} connects task inputs, execution records, and metrics.

\subsection{Task and Dataset Ecosystem}
\label{sec_task_ecosystem}

\textbf{Broad Task Suite.}
The benchmark covers 29 datasets across mathematics, knowledge, code generation, biomedical question answering, finance, and tabular reasoning. These domains combine different requirements for sequential reasoning, specialist knowledge, executable checking, and structured information integration. Dataset loaders provide task inputs, reference answers, and metric metadata to the common runner.

\textbf{G-MAS-Complex Construction.}
G-MAS-Complex contains 400 frozen multi-document tasks in four complexity tiers. Document dependencies distribute relevant information across sources, conflicting records require revision or authority resolution, and structured outputs require consistency checks. These properties combine retrieval, reconciliation, reasoning, and verification within each task. Comparisons measure how the evaluated configurations handle these coupled requirements. Tier labels describe task construction, and fixed-method scores measure empirical difficulty.

The four tiers contain 100 tasks each and share a structured answer contract. In an illustrative inventory task, D1 records 12 units, D2 revises the inventory to 9, and D3 reserves 4. The answer requires 5 available units and source identifiers D2 and D3. Selecting D1 gives 8 units, while omitting D2 produces an incomplete source record. Strict exact match checks required fields, values, rankings, references, checkpoints, and checksum constraints together. Appendix~\ref{app:complex-example} expands this example and distinguishes content errors from output-format failures.

\subsection{Algorithm Taxonomy}
\label{sec_algorithm_taxonomy}

\textbf{Definition of G-MAS.}
A graph-enhanced configuration contains at least two agent units and an explicit directed graph that constrains routing, activation, verification flow, or execution order. Classification uses the graph exposed to the adapter and the runtime decision controlled by its edges. Fixed, task-generated, and learned graphs can satisfy this criterion. Equation~\ref{eq_organization} supplies the representation used to inspect these properties.

\textbf{Single-Agent and Ordinary Multi-Agent Configurations.}
Single-agent configurations use one agent without inter-agent communication, represented here by DeepSeek and Qwen~\citep{deepseekai2025short,qwen2025short}. Ordinary multi-agent configurations coordinate agents through their dialogue or collaboration policies. The tables place the evaluated AutoGen, MADebate, and DyLAN configurations in this group~\citep{wu2023autogen,du2023multiagentdebate,liu2024dynamicllmpoweredagentnetwork}. These labels identify the reported configurations, while the adapter description specifies the organization and execution mechanism for each method.

\textbf{Graph-Enhanced Configurations.}
\looseness=-1
The selected graph-enhanced configurations cover graph construction, sparse interaction, routing, verification, and learned coordination. Table~\ref{tab:baseline-adapters} lists the corresponding methods and sources. Configuration records describe graph materialization, construction rules, runtime use, role heterogeneity, and recovery. These attributes connect method selection with the intervention protocols.

GPTSwarm optimizes prompts and graph connectivity~\citep{zhuge2024gptswarm}, while G-Designer learns a generator for task-dependent communication graphs~\citep{zhang2024gdesigner}. Configuration attributes further distinguish topology construction, runtime updates, learned components, role heterogeneity, verification, and recovery. Appendix~\ref{app:organization-analysis} defines these attributes and links graphs with execution.

\subsection{Graph Intervention Protocols}
\label{sec_graph_intervention}

Each protocol modifies one organizational factor under the shared controls of Equation~\ref{eq_shared_control}. The four protocols examine where information travels, which units process it, what they receive, and how execution continues when workers become unavailable.

\textbf{Structure Intervention.}
The structure protocol rewires directed edges while preserving the agent set, in-degree and out-degree sequences, edge count, and communication budget limit. A directed edge swap exchanges the destinations of two connections, changing communication partners while retaining local degrees. Self-loops, duplicate edges, and invalid role connections are rejected. The score change measures the response to this modification. Executed calls and messages describe how the resulting graph uses its available budget.

\textbf{Node Intervention.}
The node protocol removes selected collaboration units and their incident edges. The No critic and No specialist variants target different role groups. Expert only retains the expert nodes and their induced connections. Each variant links its available functions and communication relations with its task score.

\textbf{Information Intervention.}
The information protocol replaces eligible upstream records or intermediate candidates with format-valid incorrect candidates. The modified content reaches downstream units through the existing graph and routing rules. The corruption rate is the fraction of eligible items replaced. This protocol measures final-answer accuracy under incorrect intermediate inputs.

\textbf{Execution Intervention.}
The execution protocol disables workers during runtime and resumes from the last valid shared state through configured retry, fallback, or rerouting procedures. Increasing the failure count tests how larger disruptions affect task completion. Appendix~\ref{app:experimental-details} provides the intervention settings and shared controls.

\subsection{Evaluation Protocols}
\label{sec_evaluation_protocols}

\textbf{Utility.}
We compare single-agent, ordinary multi-agent, and graph-enhanced configurations on shared tasks. Single-agent and ordinary multi-agent results provide reference scores for assessing the utility of each configuration. These system comparisons include the complete reasoning and execution procedure. Paired interventions examine changes within a configured organization.

\textbf{Effectiveness.}
We examine task scores and method rankings across the six domains and G-MAS-Complex. Domain summaries describe performance across related tasks, while individual dataset scores capture more specific requirements. These views identify the strongest configurations and the separation among competing methods.

\textbf{Robustness.}
We compare each modified score with the same configuration's reference score. Absolute score change measures the difference in task units, and relative retention divides the modified score by a positive reference score. Rewiring ratios, unit-removal variants, corruption rates, and worker-failure counts define the intervention settings.

\textbf{Efficiency.}
We record token usage, model calls, latency, executed units and relations, and retries. For G-MAS-Complex, strict exact match describes successful completion, while token usage and latency describe execution requirements. All four dimensions use the common execution interface and task-specific scoring rules.

\section{Experiments and Analyses}
\label{sec:experiments}
\raggedbottom

Three questions organize the evaluation. \textbf{Task utility} covers Q1 to Q3. \textbf{Organizational sensitivity} covers Q4 to Q6. \textbf{Failure robustness and resource use} covers Q7 and Q8. The following experiments examine each group through specific comparisons.
\textbf{Q1 (Utility).} Which evaluated configurations improve task scores over single-agent baselines?
\textbf{Q2 (Utility).} How do the best evaluated graph-enhanced configurations compare with ordinary multi-agent systems?
\textbf{Q3 (Effectiveness).} How do existing G-MAS methods perform across different task domains?
\textbf{Q4 (Structure Intervention).} Does topology matter when the degree sequence and communication budget limits are fixed?
\textbf{Q5 (Unit Intervention).} How much do role and collaboration-unit removals affect performance?
\textbf{Q6 (Information Intervention).} How does accuracy change as the fraction of incorrect upstream items increases?
\textbf{Q7 (Execution Intervention).} How robust are G-MAS executions as the number of failed workers increases under recovery?
\textbf{Q8 (Efficiency).} How efficient are G-MAS methods?

\subsection*{\normalfont\itshape A. Experimental Setup}
\label{subsec:experimental-setup}

\looseness=-1
\textbf{Datasets}.
We evaluate seven categories. (1) Mathematics uses GSM8K and AQuA~\citep{DBLP:journals/corr/abs-2110-14168,ling-etal-2017-program}. (2) Knowledge uses MMLU-Pro and StrategyQA~\citep{wang2024mmluprorobustchallengingmultitask,geva-etal-2021-aristotle}. (3) Code uses HumanEval and LiveCodeBench~\citep{chen2021evaluatinglargelanguagemodels,jain2024livecodebenchholisticcontaminationfree}. (4) Biomedical QA uses MedQA and MedMCQA~\citep{jin2021medqa,pmlr-v174-pal22a}. (5) Finance uses ConvFinQA and FinQA~\citep{chen-etal-2022-convfinqa,chen-etal-2021-finqa}. (6) Tabular reasoning uses TabFact and WikiTableQuestions~\citep{chen2020tabfactlargescaledatasettablebased,pasupat-liang-2015-compositional}. (7) G-MAS-Complex contains 400 frozen multi-document tasks in four tiers. Full definitions are in Appendix~\ref{app:dataset-details}.

\textbf{Baselines}.
We evaluate the 17 configurations in Table~\ref{tab:full-main-results} through common adapters and evaluation controls. Appendix~\ref{app:experimental-details} describes execution, and Appendix~\ref{app:validation-protocol} defines validation records.

\textbf{Scores and variation.} The broad suite retains dataset-specific metrics, and G-MAS-Complex uses Strict EM. Table~\ref{tab:main-results} reports central scores. Appendix~\ref{app:dataset-details} separates sample-set variation, run variation, recorded estimates, and binomial standard errors. Rankings in Q1 to Q3 describe displayed central values. Intervention analyses report paired score changes and reference-score retention, while efficiency compares observed scores and resource use.

\subsection*{\normalfont\itshape B. Utility of Multi-Agent Collaboration (Q1)}
\label{subsec:q1}

To answer Q1, we compare Single-Agent, Ordinary MAS, and Graph-enhanced MAS on 13 datasets.
\vspace{-0.45\baselineskip}
\begin{table}[!ht]
\centering
\caption{Central-score comparison (\%). Bold and underlined entries mark the highest and second-highest distinct displayed scores, including ties. Appendix~\ref{app:uncertainty-sources} defines the statistical sources.}
\label{tab:main-results}
\footnotesize
\setlength{\tabcolsep}{2.7pt}
\renewcommand{\arraystretch}{1.20}
\resizebox{\textwidth}{!}{%
\begin{tabular}{l*{13}{c}}
\toprule
& \multicolumn{2}{c}{\makecell{Mathematical\\Reasoning}} & \multicolumn{2}{c}{\makecell{Knowledge \&\\Commonsense}} & \multicolumn{2}{c}{\makecell{Code\\Generation}} & \multicolumn{2}{c}{\makecell{Biomedical\\QA}} & \multicolumn{2}{c}{\makecell{Financial\\Reasoning}} & \multicolumn{2}{c}{\makecell{Tabular\\Reasoning}} & \multicolumn{1}{c}{\makecell{Complex\\Collaboration}} \\
\cmidrule(lr){2-3}\cmidrule(lr){4-5}\cmidrule(lr){6-7}\cmidrule(lr){8-9}\cmidrule(lr){10-11}\cmidrule(lr){12-13}\cmidrule(lr){14-14}
Method & GSM8K & AQuA & MMLU-Pro & StrategyQA & HumanEval & LiveCode & MedQA & MedMCQA & ConvFinQA & FinQA & TabFact & WTQ & G-MAS-Complex \\
\midrule
DeepSeek & $93.03$ & $89.89$ & \underline{$73.07$} & $84.60$ & $83.54$ & $21.90$ & $88.58$ & {\boldmath$81.27$} & $60.13$ & $38.10$ & $91.47$ & $76.73$ & \underline{$26.75$} \\
Qwen & $59.44$ & $68.90$ & $60.93$ & $79.40$ & {\boldmath$90.24$} & {\boldmath$26.86$} & $78.95$ & $73.53$ & {\boldmath$61.81$} & $28.51$ & $81.40$ & $65.27$ & $13.25$ \\
AutoGen & $92.92$ & \underline{$90.16$} & $71.80$ & $84.13$ & $81.71$ & $18.86$ & $88.74$ & $79.67$ & $60.67$ & $36.70$ & $91.67$ & $75.93$ & $23.50$ \\
MADebate & $93.20$ & $89.89$ & $71.87$ & $84.20$ & $79.88$ & $20.38$ & $89.45$ & $79.67$ & $61.01$ & $36.97$ & $91.47$ & $76.67$ & $14.00$ \\
DyLAN & $93.33$ & $89.89$ & {\boldmath$73.87$} & {\boldmath$86.67$} & $85.98$ & $22.86$ & $88.45$ & $79.93$ & $61.10$ & \underline{$38.88$} & \underline{$92.53$} & $77.40$ & $4.50$ \\
GPTSwarm & \underline{$93.66$} & {\boldmath$91.34$} & $71.20$ & $85.67$ & $85.57$ & $21.14$ & \underline{$89.63$} & \underline{$80.20$} & $61.12$ & $38.62$ & {\boldmath$92.60$} & $76.60$ & $15.50$ \\
GDesigner & \underline{$93.66$} & $89.63$ & $71.87$ & $85.00$ & $85.37$ & \underline{$23.43$} & $88.85$ & $79.00$ & \underline{$61.54$} & $38.51$ & $91.73$ & {\boldmath$80.33$} & $14.50$ \\
SparseCT & {\boldmath$93.86$} & $89.76$ & $71.33$ & $85.80$ & $85.98$ & $20.76$ & $88.92$ & $79.67$ & $60.81$ & {\boldmath$39.23$} & $91.93$ & $77.47$ & $15.25$ \\
R-GFM & $93.63$ & $90.03$ & $71.53$ & $84.40$ & $89.02$ & $21.52$ & $89.47$ & $79.33$ & $60.63$ & $38.71$ & $92.13$ & \underline{$79.80$} & $2.25$ \\
BigMAS & $93.45$ & $89.63$ & $72.20$ & $85.53$ & $85.77$ & $21.33$ & {\boldmath$89.87$} & $78.73$ & $60.69$ & $38.36$ & $92.33$ & $76.80$ & $19.25$ \\
GoAgent & $93.23$ & $89.76$ & $70.33$ & $85.27$ & $79.88$ & $21.14$ & $89.40$ & $79.13$ & $60.49$ & $37.20$ & $92.33$ & $76.20$ & {\boldmath$50.25$} \\
ARGDes & $93.48$ & $89.63$ & $71.47$ & \underline{$86.00$} & \underline{$89.63$} & $21.33$ & $88.77$ & $79.20$ & $60.20$ & $38.71$ & $91.93$ & $77.00$ & $13.25$ \\
EIB & $93.33$ & $89.37$ & $72.47$ & $85.40$ & $83.54$ & $21.33$ & $89.00$ & $79.00$ & $61.34$ & $38.54$ & $92.33$ & $79.33$ & $14.00$ \\
\bottomrule
\end{tabular}%
}
\end{table}

In Table~\ref{tab:main-results}, graph-enhanced configurations have the highest displayed central score on 7 of 13 datasets, single-agent configurations on 4, and ordinary multi-agent configurations on 2. Single-agent models have the highest central scores on both displayed code datasets and ConvFinQA. This distribution places configuration choice in the context of the target task. Our Conclusion (C1). Central-score advantages vary across tasks and configuration groups.

\subsection*{\normalfont\itshape C. Competitiveness of Graph-Enhanced MAS (Q2)}
\label{subsec:q2}

\begin{figure}[!htb]
  \centering
  \includegraphics[width=0.86\textwidth]{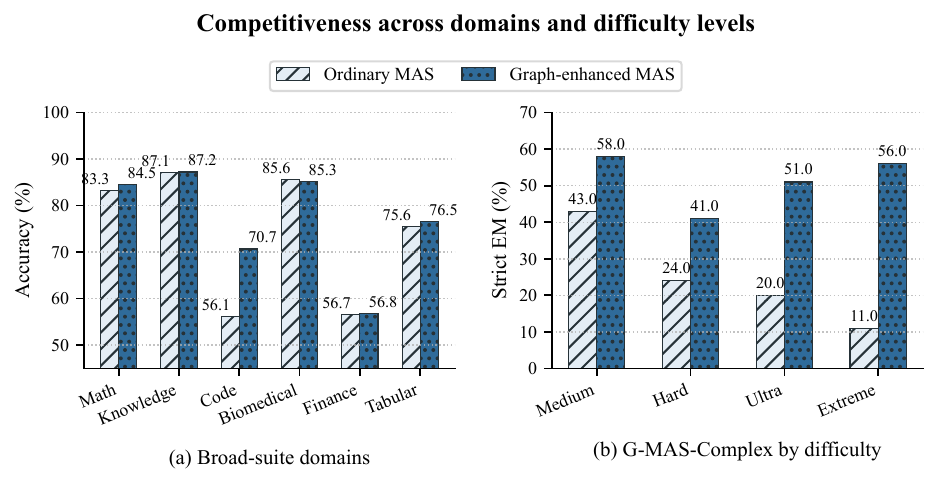}
  \caption{Group-wise maximum reported scores over evaluated Ordinary MAS and Graph-enhanced MAS configurations. Each bar selects the highest score within its group and task subset, allowing different configurations to supply different bars.}
  \label{fig:q2-competitiveness}
\end{figure}

For Q2, Figure~\ref{fig:q2-competitiveness} displays group-wise maxima over 3 ordinary and 12 graph-enhanced configurations. Each bar summarizes attainable reported scores in its group and task subset. The selected configuration can change across domains and tiers, so the comparison describes the best displayed score in each setting.

\begingroup
\lsstyle
\looseness=-2
On the four G-MAS-Complex tiers, the plotted margins are 15, 17, 31, and 45 percentage points. Across all 400 tasks, GoAgent and VeriMap~\citep{chen2026goagentgroupofagentscommunicationtopology,xu2025verificationawareplanningmultiagentsystems} score above the single-agent DeepSeek reference, while the other ten graph-enhanced configurations score below it. Tier maxima describe the best attained scores within each subset. Our Conclusion (C2). Strong complex-task performance is concentrated in a small subset of the evaluated graph-enhanced configurations.
\par
\endgroup

\subsection*{\normalfont\itshape D. Performance Across Task Domains (Q3)}
\label{subsec:q3}

\begin{figure}[!htb]
  \centering
  \includegraphics[width=0.86\textwidth]{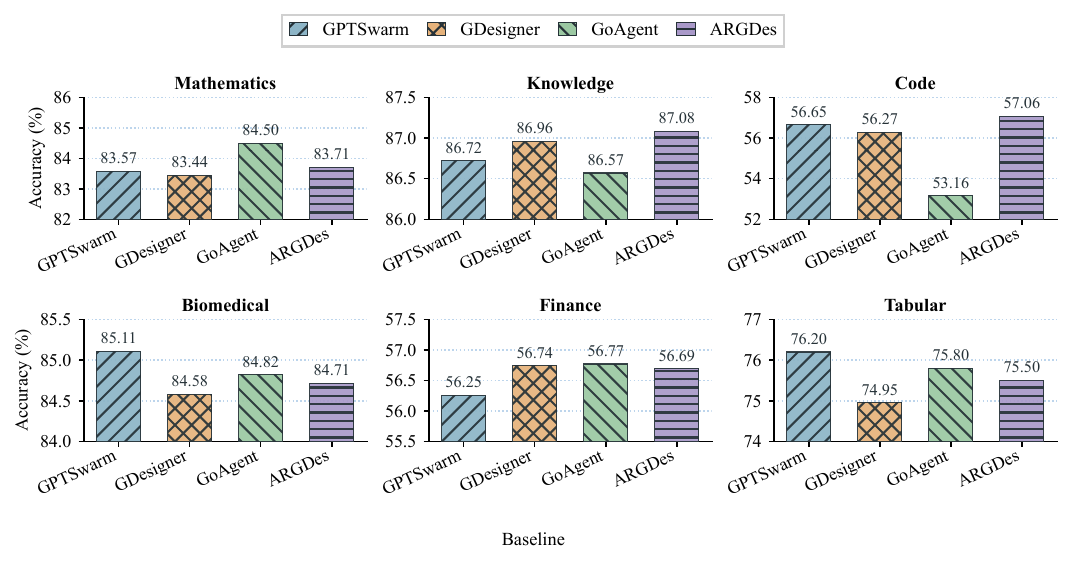}
  \caption{Domain summaries with independent panel scales. Accuracy denotes the reported aggregate score. Appendix~\ref{app:dataset-details} lists dataset-specific metrics.}
  \label{fig:q3-domain-baselines}
\end{figure}

Figure~\ref{fig:q3-domain-baselines} compares four graph-enhanced configurations across six domains.

\looseness=-1
Code ranges from 53.16\% to 57.06\%, giving the largest central-score spread of 3.90 percentage points. The spreads are 1.25 on Tabular, 1.06 on Mathematics, 0.53 on Biomedical QA, 0.52 on Finance, and 0.51 on Knowledge. Appendix~\ref{app:dataset-details} reports dataset-level variation. Our Conclusion (C3). Displayed central-score dispersion varies substantially across task domains.

\begin{figure}[!htb]
  \centering
  \includegraphics[width=\textwidth]{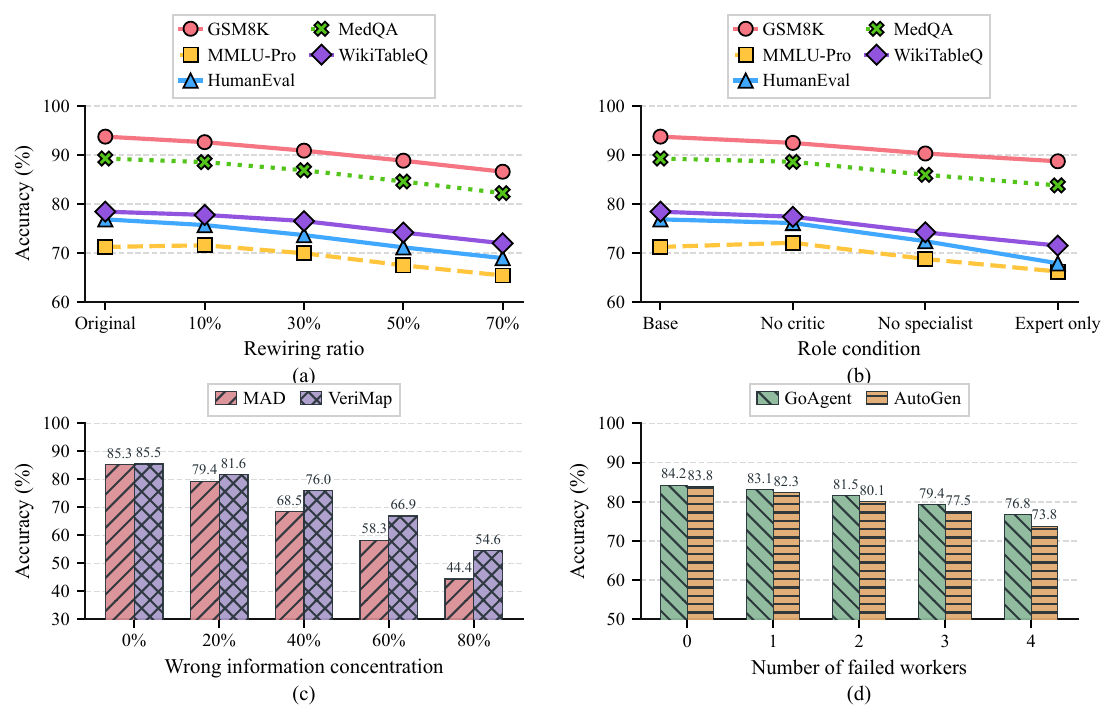}
  \caption{Responses to topology rewiring, role removal, incorrect messages, and worker failures. Panels (a) and (b) show five datasets, while panels (c) and (d) compare the displayed methods. Section~\ref{sec_graph_intervention} defines the corresponding operations.}
  \label{fig:q4-q7-interventions}
\end{figure}

\subsection*{\normalfont\itshape E. Sensitivity to Communication Structure (Q4)}
\label{subsec:q4}

\looseness=-1
We hold agents, roles, degrees, edge count, models, prompts, routing policy, and budget limits fixed. Rewiring ratios are 10\%, 30\%, 50\%, and 70\%. The five-dataset mean falls from 81.88\% to 81.21\%, 79.54\%, 77.21\%, and 74.99\%. At 70\% rewiring, the loss is 6.89 percentage points, leaving 91.59\% of the reference score. The progressive decrease associates task performance with the arrangement of communication partners in the evaluated setting. Actual calls, message counts, and context lengths characterize execution after rewiring. Our Conclusion (C4). Degree-preserving rewiring reduces scores under the specified protocol.

\subsection*{\normalfont\itshape F. Sensitivity to Collaboration-Unit Removal (Q5)}
\label{subsec:q5}

To answer Q5, we compare Base with No critic, No specialist, and Expert only. The first two remove the selected role nodes and incident edges. Expert only keeps the expert-induced subgraph. The mean falls from 81.88\% to 81.30\%, 78.30\%, and 75.59\%, respectively, giving losses of 0.58, 3.58, and 6.29 percentage points. Specialist removal incurs 6.17 times the mean loss of critic removal, while MMLU-Pro improves after critic removal. Our Conclusion (C5). Removing different collaboration units produces asymmetric score changes, with the largest loss under Expert only.

\subsection*{\normalfont\itshape G. Robustness to Incorrect Messages (Q6)}
\label{subsec:q6}

We replace eligible upstream records or candidates with format-valid incorrect candidates at corruption rates of 0\%, 20\%, 40\%, 60\%, and 80\%. The graph and execution controls remain fixed. The concentration axis in Figure~\ref{fig:q4-q7-interventions}(c) denotes this item replacement rate. MAD falls from 85.30\% to 44.41\%, and VeriMap falls from 85.50\% to 54.59\%. At 80\%, the retained scores are 52.06\% and 63.85\%, respectively. Their accuracy gap grows from 0.20 to 10.18 percentage points. Our Conclusion (C6). Similar reference scores conceal different corruption responses.

\begin{figure}[!b]
  \centering
  \includegraphics[width=0.80\textwidth]{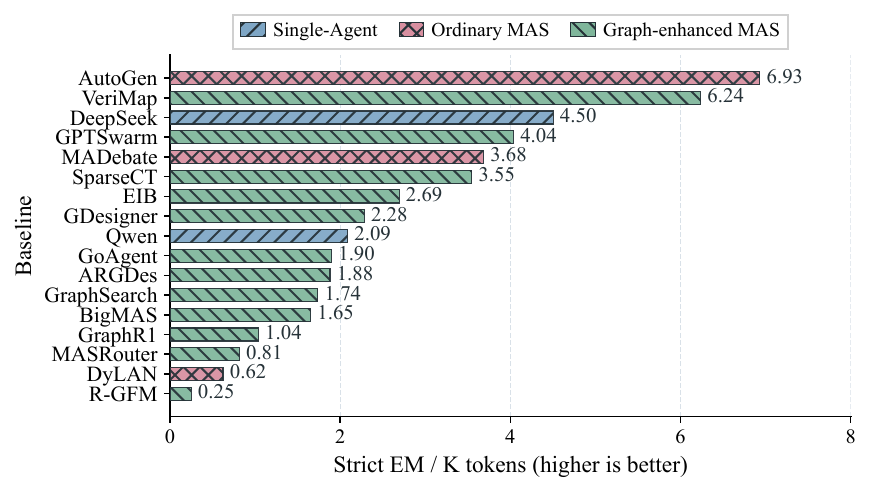}
  \caption{G-MAS-Complex efficiency. Bars show strict exact match per 1K tokens for all baselines.}
  \label{fig:q8-efficiency}
\end{figure}

\subsection*{\normalfont\itshape H. Recovery Under Worker Failures (Q7)}
\label{subsec:q7}

To answer Q7, we disable 0 to 4 workers and continue through each method's configured retry, fallback, or rerouting procedure. GoAgent decreases from 84.22\% to 76.80\%, and AutoGen decreases from 83.82\% to 73.80\%. With four failed workers, the losses are 7.42 and 10.02 percentage points, respectively. The gap grows from 0.40 to 3.00 points across the sweep. This profile compares task completion under the tested failure settings and continuation rules. Our Conclusion (C7). GoAgent retains more reference accuracy than AutoGen in this worker-failure setting.

\subsection*{\normalfont\itshape I. Efficiency (Q8)}
\label{subsec:q8}

For Q8, we compare observed operating points on G-MAS-Complex. Table~\ref{tab:q8-efficiency} places Strict EM alongside tokens, latency, and executed nodes and relations. Strict EM/K Tokens divides the strict-match percentage by mean tokens per task in thousands.

\vspace{-0.45\baselineskip}
\begin{table}[!htb]
  \centering
  \caption{Observed operating points on G-MAS-Complex. Strict EM is in percent, tokens per task in K, and latency in seconds. Nodes and edges are execution averages. Strict EM/K Tokens describes score per 1K tokens at the reported setting.}
  \label{tab:q8-efficiency}
  \footnotesize
  \setlength{\tabcolsep}{2.4pt}
  \renewcommand{\arraystretch}{1.315}
  \resizebox{\textwidth}{!}{%
  \begin{tabular}{@{}>{\raggedright\arraybackslash}p{1.85cm}*{6}{>{\centering\arraybackslash}p{1.80cm}}@{}}
    \toprule
    Method & \makecell{Strict EM\\(\%)} & \makecell{Tokens/task\\(K)} & \makecell{Strict EM/K\\Tokens} & \makecell{Latency\\(s)} & Nodes & Edges \\
    \midrule
    GoAgent & 50.25 & 26.47 & 1.90 & 45.46 & 17.25 & 67.15 \\
    VeriMap & 29.50 & 4.73 & 6.24 & 7.09 & 4.00 & 4.00 \\
    DeepSeek & 26.75 & 5.94 & 4.50 & 1.51 & 1.00 & 0.00 \\
    AutoGen & 23.50 & 3.39 & 6.93 & 6.42 & 3.00 & 2.00 \\
    BigMAS & 19.25 & 11.65 & 1.65 & 17.11 & 14.30 & 30.90 \\
    GPTSwarm & 15.50 & 3.84 & 4.04 & 6.48 & 3.95 & 2.95 \\
    SparseCT & 15.25 & 4.30 & 3.55 & 9.25 & 4.00 & 3.00 \\
    GDesigner & 14.50 & 6.35 & 2.28 & 7.95 & 6.15 & 9.05 \\
    EIB & 14.00 & 5.20 & 2.69 & 9.25 & 6.15 & 8.05 \\
    GraphSearch & 14.00 & 8.06 & 1.74 & 11.84 & 9.15 & 18.40 \\
    MADebate & 14.00 & 3.80 & 3.68 & 5.85 & 4.00 & 3.00 \\
    ARGDes & 13.25 & 7.05 & 1.88 & 8.75 & 7.15 & 11.25 \\
    Qwen & 13.25 & 6.35 & 2.09 & 6.41 & 1.00 & 0.00 \\
    GraphR1 & 10.75 & 10.35 & 1.04 & 12.60 & 11.15 & 28.55 \\
    MASRouter & 10.75 & 13.21 & 0.81 & 10.76 & 21.60 & 38.20 \\
    DyLAN & 4.50 & 7.22 & 0.62 & 12.67 & 9.15 & 12.10 \\
    R-GFM & 2.25 & 9.06 & 0.25 & 21.27 & 10.15 & 13.10 \\
    \bottomrule
  \end{tabular}%
  }
\end{table}

Figure~\ref{fig:q8-efficiency} ranks observed score-per-token values. AutoGen and VeriMap have the highest ratios, while GoAgent has the highest Strict EM. Compared with DeepSeek, GoAgent gains 23.50 percentage points while using 4.46 times the tokens and 30.11 times the latency. Tokens describe model usage, and latency describes elapsed execution under the associated deployment settings. Our Conclusion (C8). The highest accuracy and highest score per token occur at different operating points.

\begingroup
\setlength{\parskip}{0.5pc plus 2pt}
\section{Conclusion and Future Directions}
\label{sec:conclusion}

OpenMAS-GCom evaluates how graph-enhanced multi-agent systems respond to organizational changes. Its executable representation separates collaboration units, communication relations, shared information, and execution policy. Four paired intervention families modify structure, unit composition, intermediate messages, and worker availability under shared controls. The broad task suite and G-MAS-Complex connect these responses with task outcomes and resource use. The shared interface associates each comparison with its organization settings, intervention parameters, task scores, and execution records.

Three findings organize the evaluated results. First, central-score advantages vary across tasks, and GoAgent and VeriMap are the two graph-enhanced configurations above DeepSeek on G-MAS-Complex. Second, degree-preserving rewiring and collaboration-unit removal produce unequal score changes under their respective settings. At 70\% rewiring, the mean score declines by 6.89 points, while specialist removal produces a larger mean decline than critic removal. Third, similar reference scores accompany different responses to incorrect messages and unavailable workers. MAD and VeriMap separate by 10.18 points at 80\% corruption, and the accuracy gap between GoAgent and AutoGen increases with worker failures. The configuration with the highest Strict EM also differs from the configuration with the highest score per token.

These findings characterize the tested configurations, tasks, and disruption settings. Paired score changes summarize the response of the complete modified execution. G-MAS-Complex targets document dependencies, record conflicts, and structured completion, while the broad suite retains each dataset's task-specific metric. Resource comparisons describe observed execution settings through scores, token usage, and latency. These views guide configuration selection and organizational diagnosis. Extensions to longer-horizon, interactive, and multimodal tasks can use the same organization interface and paired comparison protocol.

\par
\clearpage
\endgroup
\setlength{\parskip}{0.5pc}
\appendix
\section{Appendix}

This appendix describes datasets, baseline adapters, metrics, interventions, evaluation strategies, execution metadata, parameter settings, supplementary results, and reproducibility records.

\subsection{Dataset Description}
\label{app:dataset-details}

OpenMAS-GCom contains a broad task suite and a separately constructed G-MAS-Complex stress suite. The broad suite contains 29 datasets across six domains. Mathematics evaluates arithmetic, symbolic, and competition-style reasoning with GSM8K, SVAMP, MultiArith, AQuA, ASDiv-A, Game-of-24, and MATH~\citep{DBLP:journals/corr/abs-2110-14168,patel-etal-2021-nlp,roy-roth-2015-solving,ling-etal-2017-program,miao-etal-2020-diverse,yao2023treethoughtsdeliberateproblem,DBLP:journals/corr/abs-2103-03874}. Knowledge and commonsense uses MMLU, MMLU-Redux, MMLU-Pro, StrategyQA, CommonsenseQA, ARC-Easy, and ARC-Challenge to test subject knowledge, robust multiple-choice answering, implicit reasoning, and science QA~\citep{hendrycks2021measuringmassivemultitasklanguage,gema2025mmlu,wang2024mmluprorobustchallengingmultitask,geva-etal-2021-aristotle,talmor-etal-2019-commonsenseqa,clark2018thinksolvedquestionanswering}. Code uses HumanEval, HumanEval++, LiveCodeBench-v6, and MultiAgentBench-Coding for executable program synthesis and collaborative coding evaluation~\citep{chen2021evaluatinglargelanguagemodels,liu2023codegeneratedchatgptreally,jain2024livecodebenchholisticcontaminationfree,zhu2025multiagentbench}. Biomedical tasks include PubMedQA, MedQA, MedMCQA, and the medical subset of MMLU~\citep{jin-etal-2019-pubmedqa,jin2021medqa,pmlr-v174-pal22a,hendrycks2021measuringmassivemultitasklanguage}. Finance uses TAT-QA, ConvFinQA, and FinQA to evaluate numerical reasoning over financial text and tables~\citep{zhu-etal-2021-tat,chen-etal-2022-convfinqa,chen-etal-2021-finqa}. Tabular reasoning uses TabFact, WikiSQL, WikiTableQuestions, and SQA for table verification, semantic parsing, and sequential table QA~\citep{chen2020tabfactlargescaledatasettablebased,zhong2017seq2sqlgeneratingstructuredqueries,pasupat-liang-2015-compositional,iyyer2017search}.

\begin{table}[!htbp]
\centering
\caption{\textsc{Broad-suite dataset summary.} Domain-level statistics summarize the broad suite. Per-dataset sample counts are shown in Table~\ref{tab:full-main-results}.}
\label{tab:dataset-domain-summary}
\footnotesize
\setlength{\tabcolsep}{4pt}
\renewcommand{\arraystretch}{1.08}
\begin{tabular}{@{}L{1.65cm}c L{3.5cm} L{4.8cm}@{}}
\toprule
Domain & \#Datasets & Datasets & Primary evaluation target \\
\midrule
Mathematics & 7 & GSM8K, SVAMP, MultiArith, AQuA, ASDiv-A, Game-of-24, MATH & Numeric answer, multiple-choice answer, valid expression, or competition-style final answer \\
Knowledge and commonsense & 7 & MMLU, MMLU-Redux, MMLU-Pro, StrategyQA, CommonsenseQA, ARC-Easy, ARC-Challenge & Multiple-choice or yes/no answer after robust answer extraction \\
Code & 4 & HumanEval, HumanEval++, LiveCodeBench-v6, MultiAgentBench-Coding & Unit-test pass rate, executable correctness, or code-task quality score \\
Biomedical & 4 & PubMedQA, MedQA, MedMCQA, MMLU-Med & Biomedical yes/no/maybe answer or medical multiple-choice answer \\
Finance & 3 & TAT-QA, ConvFinQA, FinQA & Normalized numerical or textual financial answer \\
Tabular & 4 & TabFact, WikiSQL, WikiTableQuestions, SQA & Table entailment, table-query answer, or sequential table answer \\
\bottomrule
\end{tabular}
\end{table}

\textbf{G-MAS-Complex.} We additionally construct 400 frozen multi-document tasks to stress collaboration rather than isolated recall. Each task contains a query, source documents, dependency metadata, conflict annotations, and a machine-checkable JSON output contract. The documents have overlapping contents and distinct functions. Some provide authoritative revisions, some provide intermediate records, some are distractors, and some define constraints that must be combined before the final answer can be produced. This design makes the suite sensitive to delegation, provenance tracking, conflict resolution, and final verification.

\begin{table}[!htbp]
\centering
\caption{\textsc{G-MAS-Complex tier construction.} Each tier contains 100 frozen tasks.}
\label{tab:gmas-complex-tiers}
\footnotesize
\setlength{\tabcolsep}{3.6pt}
\renewcommand{\arraystretch}{1.08}
\resizebox{\textwidth}{!}{%
\begin{tabular}{@{}L{1.35cm}L{3.0cm}L{5.0cm}L{4.15cm}@{}}
\toprule
Tier & Main stress factor & Document and conflict pattern & Required behavior \\
\midrule
Medium & Multi-document lookup & Several documents with local dependencies and light distractors & Retrieve relevant records and return a structured answer \\
Hard & Cross-document dependency & Chained records with competing revisions or aliases & Resolve dependencies before producing fields and references \\
Ultra & Conflict and provenance & Multiple plausible records, authority rules, and source checkpoints & Select authoritative records and justify structured fields \\
Extreme & End-to-end audit & Dense dependencies, distractors, revision conflicts, and checksum constraints & Combine retrieval, conflict resolution, ranking, and verification \\
\bottomrule
\end{tabular}
}
\end{table}

For each task, the target answer is represented as
\begin{equation}
z_i=(f_i,v_i,r_i,\ell_i,h_i,\chi_i),
\end{equation}
where $f_i$ contains required fields, $v_i$ contains normalized values, $r_i$ contains ranking or ordering requirements, $\ell_i$ contains source references, $h_i$ contains checkpoint records, and $\chi_i$ contains checksum-style consistency constraints. The output parser first checks whether the response can be mapped to the requested JSON schema, then normalizes scalar values, lists, and source identifiers. A task is counted as correct only when every required component satisfies the contract.

\textbf{Complex-task validation example.}
\label{app:complex-example-overview}
Each stress-suite instance can also be written as $x_i=(q_i,\mathcal{D}_i,\Gamma_i,\Omega_i)$, where $q_i$ is the task query, $\mathcal{D}_i$ is the source-document collection, $\Gamma_i$ specifies dependencies and record conflicts, and $\Omega_i$ defines the output contract.
For example, an inventory task may contain three records. $D_1$ states that 12 units are available, $D_2$ revises the inventory to 9 units, and $D_3$ reserves 4 units.
The correct answer is therefore 5 available units with source identifiers $D_2$ and $D_3$.
Using $D_1$ instead of $D_2$ yields a content error, while returning the right value without the required source identifiers yields a contract error.
The strict scorer rejects both cases because it checks field values, provenance, and schema validity jointly.

\subsection{Baseline Description}
\label{app:experimental-details}

Every method is wrapped as an executable organization $\mathcal{O}=(V,E,S,\pi)$ before evaluation. This adapter layer separates method-specific collaboration behavior from shared benchmark components such as dataset loading, backend calls, answer parsing, metric calculation, trace logging, and cost accounting. Single-agent baselines contain one runnable unit and no communication edges. Ordinary MAS baselines contain multiple units with predefined collaboration patterns. Graph-enhanced MAS baselines expose organization graphs, learned or searched routing, sparse communication, verification links, or recovery policies through the same runner interface.

\begin{table}[!t]
\centering
\caption{\textsc{Baseline families and organization adapters.}}
\label{tab:baseline-adapters}
\footnotesize
\setlength{\tabcolsep}{3.2pt}
\renewcommand{\arraystretch}{1.08}
\begin{tabular}{@{}L{2.0cm}L{4.0cm}L{5.0cm}@{}}
\toprule
Family & Methods & Adapter representation \\
\midrule
Single-agent & DeepSeek, Qwen~\citep{deepseekai2025short,qwen2025short} & One runnable unit, empty edge set, direct answer parser \\
Ordinary MAS & AutoGen, MADebate, DyLAN~\citep{wu2023autogen,du2023multiagentdebate,liu2024dynamicllmpoweredagentnetwork} & Fixed role set with predefined debate, refinement, or dynamic selection policy \\
Graph construction and routing & GPTSwarm, GDesigner, GraphSearch, GraphR1, MASRouter~\citep{zhuge2024gptswarm,zhang2024gdesigner,liu2026graphsearchagenticsearchaugmentedreasoning,luo2026graphr1agenticgraphragframework,yue2025masrouterlearningroutellms} & Explicit graph or search/routing module mapped to typed directed relations \\
Sparse or role-specialized G-MAS & SparseCT, R-GFM, BigMAS, ARGDes, EIB~\citep{li-etal-2024-improving-multi,liu2026learninggraphfoundationmodels,hao2026braininspiredgraphmultiagentsystems,li2026assemble,shen2025eib} & Specialist units and sparse communication structures mapped to runnable nodes and edges \\
Verification and recovery-oriented G-MAS & GoAgent, VeriMap~\citep{chen2026goagentgroupofagentscommunicationtopology,xu2025verificationawareplanningmultiagentsystems} & Verification, fallback, or rerouting policies exposed as execution-policy components \\
\bottomrule
\end{tabular}
\end{table}

\textbf{Organization configuration and execution.}
\label{app:organization-analysis-overview}
The configured organization and the realized execution trace are recorded separately.
The organization specifies available units, roles, communication relations, routing policy, and recovery policy, while the trace records which units actually execute for a particular input.
Execution can differ from the static configuration as messages arrive, candidate answers are rejected, workers become unavailable, or recovery procedures are triggered.
This distinction separates available units from executed units and connects the graph specified by a method with the communication actually used to produce a prediction.

\FloatBarrier
\subsection{Metric Description}
\label{app:metric-description}

OpenMAS-GCom reports the official or commonly adopted metric for each dataset whenever possible. For classification and question-answering datasets, the parser maps free-form generations to the answer space before applying exact agreement with the reference. For code tasks, generated programs are evaluated with the corresponding pass@1-style executable test protocol. For tabular and financial reasoning, answers are normalized before comparison, including number formatting, option markers, and short textual aliases when supported by the original benchmark metric.

For a dataset $\mathcal{D}=\{(x_i,y_i)\}_{i=1}^{n}$ and method $F$, the broad-suite score is
\begin{equation}
s(F,\mathcal{D})=\frac{1}{n}\sum_{i=1}^{n} m(\operatorname{parse}(F(x_i)),y_i),
\end{equation}
\looseness=-1
where $m$ is the dataset-specific metric. For QA datasets, $m$ is accuracy or exact match. For code datasets, $m$ is executable correctness. For MultiAgentBench-Coding, $m$ is CodeQ.

For G-MAS-Complex, strict exact match is
\begin{equation}
\operatorname{StrictEM}(\hat{z}_i,z_i)=
\mathbf{1}[\hat{z}_i\models\Omega_i \land \hat{z}_i=z_i],
\end{equation}
and the suite score is
\begin{equation}
s_{\mathrm{complex}}(F)=\frac{1}{400}\sum_{i=1}^{400}\operatorname{StrictEM}(\hat{z}_i,z_i).
\end{equation}
A prediction is correct only when all required fields, values, rankings, source references, checkpoints, and checksum constraints are satisfied. Partially correct answers remain in the raw prediction logs but receive zero strict exact-match credit.

\looseness=-1
Efficiency is reported with the same trace fields for all methods. We record token usage, model calls, executed nodes, executed edges, retries, latency, and available cost statistics. For G-MAS-Complex, Table~\ref{tab:q8-efficiency} additionally reports $\mathrm{Strict\ EM}/\mathrm{K\ Tokens}$, the strict-match percentage per 1K tokens.

\subsection{Robustness Intervention Description}
\label{app:intervention-description}

For every intervention, the task list, backend configuration, prompt template, answer parser, metric, routing policy, and runtime budget remain fixed unless the intervention explicitly targets the corresponding organization component. Let
\begin{equation}
\mathcal{C}=\{\mathcal{D}_{\mathcal{T}},\Theta,M_{\mathcal{T}},P\}
\end{equation}
denote the shared controls. For intervention type $r$, we construct
\begin{equation}
\mathcal{O}'_{r}=\mathcal{I}_{r}(\mathcal{O},\Delta\mathcal{O}_{r}),
\qquad
r\in\{\mathrm{str},\mathrm{node},\mathrm{info},\mathrm{exec}\},
\end{equation}
and measure
\begin{equation}
\Delta_{r}(F,\mathcal{T})=
s(F,\mathcal{O}'_{r},\mathcal{T})-s(F,\mathcal{O},\mathcal{T}).
\end{equation}
\looseness=-1
A negative value indicates lower performance after modification under the shared controls.

\begin{table}[!htbp]
\centering
\caption{\textsc{Controls preserved by diagnostic interventions.} Yes means the factor is kept fixed relative to the reference run.}
\label{tab:intervention-controls}
\footnotesize
\setlength{\tabcolsep}{3pt}
\renewcommand{\arraystretch}{1.08}
\begin{tabular}{@{}L{2.8cm}cccccccc@{}}
\toprule
Intervention & Samples & Backend & Prompts & Parser & Budget & Roles & Edges & Degree seq. \\
\midrule
Structure rewiring & Yes & Yes & Yes & Yes & Yes & Yes & Changed & Yes \\
Node removal & Yes & Yes & Yes & Yes & Yes & Changed & Changed & No \\
Information injection & Yes & Yes & Yes & Yes & Yes & Yes & Yes & Yes \\
Execution failure & Yes & Yes & Yes & Yes & Yes & Yes & Yes & Yes \\
\bottomrule
\end{tabular}
\end{table}

\textbf{Structure intervention.} We evaluate the Original graph and degree-preserving rewiring ratios of 10\%, 30\%, 50\%, and 70\%. Directed edge pairs $u\!\to\!a$ and $v\!\to\!b$ are proposed for the swap $u\!\to\!b$ and $v\!\to\!a$. Proposals that create self-loops, duplicate edges, or invalid role connections are rejected. The actual ratio and random seed are recorded.

\textbf{Node intervention.} The No critic and No specialist variants remove the corresponding role nodes and all incident edges. The Expert only variant retains expert nodes and their induced communication edges. This intervention changes collaboration-unit composition while keeping the task, model, prompt, parser, routing, and budget controls fixed.

\textbf{Information intervention.} For concentration $c\in\{0,.2,.4,.6,.8\}$, a fraction $c$ of eligible upstream records or candidate items is replaced with a format-valid but incorrect candidate and forwarded through the unchanged downstream graph. We report task accuracy and whether final answers follow the injected candidate. The trace further records whether the injected candidate reaches downstream units and whether it is adopted, corrected, or ignored before the final answer.

\looseness=-1
\textbf{Execution intervention.} For each failure count $k\in\{0,1,2,3,4\}$, $k$ workers are marked unavailable at selected execution points. The last valid shared state is recorded and the method resumes through its configured retry, fallback, or rerouting mechanism. $k=0$ denotes no failure.

\subsection{Evaluation Strategies}
\label{app:evaluation-strategies}

\looseness=-1
The benchmark follows four evaluation perspectives, matching the main questions in Section~\ref{sec:experiments}. Necessity compares single-agent, ordinary MAS, and graph-enhanced MAS under identical task and metric settings. Effectiveness compares method behavior across the six broad-suite domains and the four G-MAS-Complex tiers. Robustness applies the four controlled interventions above to isolate communication topology, role composition, information quality, and execution recovery. Efficiency reports quality together with token usage, model calls, latency, executed nodes and edges, and retries.

\begin{table}[!htbp]
\centering
\caption{\textsc{Evaluation strategy summary.}}
\label{tab:evaluation-strategies}
\footnotesize
\setlength{\tabcolsep}{4pt}
\renewcommand{\arraystretch}{1.08}
\begin{tabular}{@{}L{1.7cm}L{4.0cm}L{5.3cm}@{}}
\toprule
Perspective & Main comparison & Reported results \\
\midrule
Necessity & Single-agent vs. ordinary MAS vs. graph-enhanced MAS & Main 13-dataset comparison and full 29-dataset table \\
Effectiveness & Method behavior across domains and complex-task tiers & Domain averages, G-MAS-Complex tiers, and complete method rankings \\
Robustness & Structure, node, information, and execution interventions & Performance changes under controlled organizational perturbations \\
Efficiency & Task quality relative to execution cost & Strict EM, tokens, latency, executed nodes, executed edges, and Strict EM/K tokens \\
\bottomrule
\end{tabular}
\end{table}

\subsection{Experiment Environment}
\label{app:experiment-environment}

The execution record associates each run with its task, organization, backend configuration, and resource measurements. All methods are executed through the same runner interface. Each run records the dataset identifier, sample identifier, backend configuration, prompt-template hash, organization hash, final prediction, parsed prediction, metric result, token usage, model calls, executed nodes and edges, retries, latency, and termination status. Intervention runs additionally record the reference organization hash, intervention type, intervention parameters, changed nodes or edges when applicable, injected-record identifiers, failed-worker identifiers, and selected recovery action.

\begin{table}[!htbp]
\centering
\caption{\textsc{Recorded environment and run metadata.} The record links task, organization, backend, prediction, cost, and intervention fields.}
\label{tab:environment-metadata}
\footnotesize
\setlength{\tabcolsep}{4pt}
\renewcommand{\arraystretch}{1.08}
\begin{tabular}{@{}L{2.5cm}L{8.8cm}@{}}
\toprule
Category & Recorded fields \\
\midrule
Task state & Dataset id, sample id, split or frozen-list identifier, gold answer, parser version \\
Organization state & Method id, organization hash, node list, edge list, role configuration, routing policy \\
Backend state & Backend configuration id, prompt-template hash, runtime limits, random seed when used \\
Prediction state & Raw response, parsed prediction, invalid-parse flag, metric result, final answer \\
Cost state & Token usage, model calls, executed nodes, executed edges, retries, latency, termination reason \\
Intervention state & Reference organization hash, intervention type, changed nodes/edges, injected records, failed workers, recovery action \\
\bottomrule
\end{tabular}
\end{table}

Before aggregation, the audit script checks that reference and intervention runs share the same task identifiers, parser version, backend configuration, prompt template, and stopping budget. Runs that fail these consistency checks are excluded from aggregate tables until regenerated.

\textbf{Validation protocol.}
\label{app:validation-protocol-overview}
Each run produces a validation record before it enters an aggregate table.
The record includes the dataset and sample identifiers, method configuration, parser version, raw answer, parsed answer, metric value, and trace hash.
For G-MAS-Complex, the validator additionally distinguishes output-format failures from content failures. Format failures cannot be mapped to $\Omega_i$, while content failures satisfy the schema but miss required values, rankings, references, checkpoints, or checksum constraints.
This separation keeps structured-output reliability visible instead of folding every failure into a single exact-match score.

\subsection{Hyperparameter Settings}
\label{app:hyperparameter-settings}

All comparisons freeze task samples, prompt templates, answer parsers, backend configuration, routing policy, and runtime limits within a comparison track. The broad suite follows fixed dataset-specific evaluation lists. Datasets with repeated-sampling analysis use three matched 500-instance samples where applicable. G-MAS-Complex is fixed. Every method is evaluated on the same 400 frozen tasks. Intervention sweeps use the discrete settings in Table~\ref{tab:hyperparameter-settings}. No intervention changes the task text, gold answer, scoring parser, or input ordering.

\begin{table}[!htbp]
\centering
\caption{\textsc{Benchmark-level hyperparameter and sweep settings.}}
\label{tab:hyperparameter-settings}
\footnotesize
\setlength{\tabcolsep}{4pt}
\renewcommand{\arraystretch}{1.08}
\begin{tabular}{@{}L{2.6cm}L{8.7cm}@{}}
\toprule
Component & Setting \\
\midrule
Broad-suite sampling & Full evaluation split where used, otherwise fixed ordered subset, with three matched 500-instance samples for repeated-sampling datasets \\
G-MAS-Complex sampling & 400 frozen tasks, 100 tasks each for Medium, Hard, Ultra, and Extreme \\
Structure rewiring & Original, 10\%, 30\%, 50\%, and 70\% degree-preserving rewiring \\
Node intervention & Base, No critic, No specialist, Expert only \\
Information intervention & Incorrect-information concentration $c\in\{0,.2,.4,.6,.8\}$ \\
Execution intervention & Failed workers $k\in\{0,1,2,3,4\}$ \\
Uncertainty reporting & Three-sample standard deviation where available, otherwise run-seed standard deviation, recorded estimate, or binomial standard-error proxy \\
\bottomrule
\end{tabular}
\end{table}

\subsection{Uncertainty Sources}
\label{app:uncertainty-sources}

The detailed tables distinguish five statistical sources through cell-level codes. Table~\ref{tab:uncertainty-sources} records the quantity attached to each score and the available repetition information. Standard deviations, analytic standard errors, and empirical estimates retain their distinct statistical meanings. Ranking markers order the displayed central scores, including equal numerical values. A paired score comparison specifies the common statistic and the matched task and seed records.

\begin{table}[H]
\centering
\footnotesize
\setlength{\tabcolsep}{3.5pt}
\renewcommand{\arraystretch}{1.18}
\begin{tabular}{@{}C{0.055\textwidth}C{0.065\textwidth}L{0.585\textwidth}L{0.22\textwidth}@{}}
\toprule
Code & Cells & Recorded quantity & Repetition information \\
\midrule
$\mathrm{s}$ & 221 & Mean and sample standard deviation across sampled task sets. Each set contains 500 instances. & Three sample sets \\
$\mathrm{r}$ & 152 & Fixed reference score with an attached run-seed standard deviation. & Run count unrecorded \\
$\mathrm{e}$ & 34 & Reference score with a recorded empirical estimate combining dataset and method components at weights 0.70 and 0.30. & Component counts unrecorded \\
$\mathrm{b}$ & 100 & Reference score with a binomial standard-error estimate computed from that score and the displayed evaluation count $N$. & Analytic estimate \\
$\mathrm{m}$ & 3 & Backup run-seed mean and standard deviation for cells with an empty reference metric. & Run count unrecorded \\
\bottomrule
\end{tabular}
\caption{Statistical sources and cell counts for the 510 entries in the detailed result tables.}
\label{tab:uncertainty-sources}
\end{table}

For the three percentage scores $y_1,y_2,y_3$, code $\mathrm{s}$ uses
\begin{equation}
\bar{y}=\frac{1}{3}\sum_{j=1}^{3}y_j,\qquad
s=\sqrt{\frac{1}{2}\sum_{j=1}^{3}(y_j-\bar{y})^2}.
\end{equation}
For code $\mathrm{b}$, let $p$ be the reported score divided by 100. The recorded estimate in percentage points is $100\sqrt{p(1-p)/N}$, using an independent binary-outcome model. Applying this model to CodeQ requires a binary task-outcome definition in the scoring record. The three code $\mathrm{m}$ entries are DeepSeek, BigMAS, and MASRouter on MultiAgentBench-Coding. Source records preserve the central value and the statistical quantity as separate fields.

\subsection{Supplementary Results}
\label{app:supplementary-results}

The complete 29-dataset and 17-baseline comparison is reported in Table~\ref{tab:full-main-results}. The repeated-sampling summary for the 13 matched-sampling datasets is provided in Table~\ref{tab:mean-std-results}. These tables are separated from the dataset description to keep the appendix layout stable.

\FloatBarrier
\noindent\textbf{Reading the detailed results.} Each entry reports a central score in percent, followed by an uncertainty term in percentage points and a source code. Code $\mathrm{s}$ gives the mean and sample standard deviation across three 500-instance sample sets. Code $\mathrm{r}$ appends a recorded run-seed standard deviation to the fixed reference score. Code $\mathrm{e}$ appends the recorded estimate combining dataset and method components with weights 0.70 and 0.30. Code $\mathrm{b}$ appends a binomial standard-error estimate using the displayed $N$. Code $\mathrm{m}$ gives a backup run-seed mean and standard deviation for the three cells whose reference metric is missing. Bold and underlined scores mark the highest and second-highest distinct values within each dataset, including ties. Horizontal rules separate single-agent, ordinary multi-agent, and graph-enhanced configurations.

\begin{table}[H]
\centering
\footnotesize
\setlength{\tabcolsep}{2.0pt}
\renewcommand{\arraystretch}{1.10}
\begin{tabular}{lccccccc}
\toprule
Method & \makecell{GSM8K\\Acc.\\$N=1319$} & \makecell{SVAMP\\Acc.\\$N=1000$} & \makecell{MultiArith\\Acc.\\$N=600$} & \makecell{AQuA\\Acc.\\$N=254$} & \makecell{ASDiv-A\\Acc.\\$N=500$} & \makecell{Game-of-24\\Acc.\\$N=1362$} & \makecell{MATH\\Acc.\\$N=500$}\\
\midrule
DeepSeek & \makecell{$\underline{93.71}$\\$(0.33)^{\mathrm{r}}$} & \makecell{$94.00$\\$(0.10)^{\mathrm{r}}$} & \makecell{$98.50$\\$(0.25)^{\mathrm{r}}$} & \makecell{$89.76$\\$(0.45)^{\mathrm{r}}$} & \makecell{$88.53$\\$(1.10)^{\mathrm{s}}$} & \makecell{$58.88$\\$(1.33)^{\mathrm{b}}$} & \makecell{$55.73$\\$(3.64)^{\mathrm{s}}$}\\
Qwen & \makecell{$59.44$\\$(0.56)^{\mathrm{e}}$} & \makecell{$90.90$\\$(0.91)^{\mathrm{b}}$} & \makecell{$97.17$\\$(0.68)^{\mathrm{b}}$} & \makecell{$68.90$\\$(0.86)^{\mathrm{e}}$} & \makecell{$86.87$\\$(0.42)^{\mathrm{s}}$} & \makecell{$21.00$\\$(1.10)^{\mathrm{b}}$} & \makecell{$45.20$\\$(2.27)^{\mathrm{s}}$}\\
\midrule
AutoGen & \makecell{$92.87$\\$(0.34)^{\mathrm{r}}$} & \makecell{$93.70$\\$(0.77)^{\mathrm{b}}$} & \makecell{$98.50$\\$(0.19)^{\mathrm{r}}$} & \makecell{$88.98$\\$(1.72)^{\mathrm{r}}$} & \makecell{$88.53$\\$(0.99)^{\mathrm{s}}$} & \makecell{$\underline{67.91}$\\$(1.26)^{\mathrm{b}}$} & \makecell{$56.13$\\$(2.66)^{\mathrm{s}}$}\\
MADebate & \makecell{$92.87$\\$(0.24)^{\mathrm{r}}$} & \makecell{$94.20$\\$(0.74)^{\mathrm{b}}$} & \makecell{$\mathbf{98.83}$\\$(0.00)^{\mathrm{r}}$} & \makecell{$89.37$\\$(0.23)^{\mathrm{r}}$} & \makecell{$88.53$\\$(0.99)^{\mathrm{s}}$} & \makecell{$62.70$\\$(1.31)^{\mathrm{b}}$} & \makecell{$55.27$\\$(3.42)^{\mathrm{s}}$}\\
DyLAN & \makecell{$93.33$\\$(0.41)^{\mathrm{e}}$} & \makecell{$94.00$\\$(0.25)^{\mathrm{r}}$} & \makecell{$\underline{98.67}$\\$(0.29)^{\mathrm{r}}$} & \makecell{$89.76$\\$(0.99)^{\mathrm{r}}$} & \makecell{$\mathbf{88.80}$\\$(1.40)^{\mathrm{s}}$} & \makecell{$62.41$\\$(1.31)^{\mathrm{b}}$} & \makecell{$56.07$\\$(3.24)^{\mathrm{s}}$}\\
\midrule
GPTSwarm & \makecell{$93.56$\\$(0.43)^{\mathrm{r}}$} & \makecell{$94.10$\\$(0.46)^{\mathrm{r}}$} & \makecell{$98.50$\\$(0.25)^{\mathrm{r}}$} & \makecell{$\mathbf{91.34}$\\$(0.68)^{\mathrm{e}}$} & \makecell{$88.67$\\$(0.81)^{\mathrm{s}}$} & \makecell{$65.71$\\$(0.85)^{\mathrm{r}}$} & \makecell{$57.00$\\$(2.84)^{\mathrm{s}}$}\\
GDesigner & \makecell{$93.63$\\$(0.27)^{\mathrm{r}}$} & \makecell{$94.00$\\$(0.06)^{\mathrm{r}}$} & \makecell{$\mathbf{98.83}$\\$(0.10)^{\mathrm{r}}$} & \makecell{$88.98$\\$(0.82)^{\mathrm{r}}$} & \makecell{$88.60$\\$(1.31)^{\mathrm{s}}$} & \makecell{$67.62$\\$(1.27)^{\mathrm{b}}$} & \makecell{$56.73$\\$(3.37)^{\mathrm{s}}$}\\
SparseCT & \makecell{$\mathbf{93.86}$\\$(0.41)^{\mathrm{e}}$} & \makecell{$93.90$\\$(0.76)^{\mathrm{b}}$} & \makecell{$\underline{98.67}$\\$(0.10)^{\mathrm{r}}$} & \makecell{$89.76$\\$(0.68)^{\mathrm{r}}$} & \makecell{$\mathbf{88.80}$\\$(0.92)^{\mathrm{s}}$} & \makecell{$64.54$\\$(1.30)^{\mathrm{b}}$} & \makecell{$\underline{57.13}$\\$(3.87)^{\mathrm{s}}$}\\
GraphSearch & \makecell{$93.56$\\$(0.13)^{\mathrm{r}}$} & \makecell{$94.10$\\$(0.21)^{\mathrm{r}}$} & \makecell{$\underline{98.67}$\\$(0.00)^{\mathrm{r}}$} & \makecell{$89.37$\\$(1.20)^{\mathrm{r}}$} & \makecell{$88.67$\\$(0.90)^{\mathrm{s}}$} & \makecell{$67.55$\\$(1.27)^{\mathrm{b}}$} & \makecell{$56.40$\\$(3.46)^{\mathrm{s}}$}\\
GraphR1 & \makecell{$92.87$\\$(0.27)^{\mathrm{r}}$} & \makecell{$93.60$\\$(0.55)^{\mathrm{r}}$} & \makecell{$98.50$\\$(0.10)^{\mathrm{r}}$} & \makecell{$\underline{90.55}$\\$(0.23)^{\mathrm{r}}$} & \makecell{$88.60$\\$(1.06)^{\mathrm{s}}$} & \makecell{$66.89$\\$(1.28)^{\mathrm{b}}$} & \makecell{$56.73$\\$(2.72)^{\mathrm{s}}$}\\
R-GFM & \makecell{$93.18$\\$(0.20)^{\mathrm{r}}$} & \makecell{$93.80$\\$(0.26)^{\mathrm{r}}$} & \makecell{$98.50$\\$(0.10)^{\mathrm{r}}$} & \makecell{$89.76$\\$(0.91)^{\mathrm{r}}$} & \makecell{$\mathbf{88.80}$\\$(1.44)^{\mathrm{s}}$} & \makecell{$66.08$\\$(1.28)^{\mathrm{b}}$} & \makecell{$55.73$\\$(3.88)^{\mathrm{s}}$}\\
BigMAS & \makecell{$93.18$\\$(0.19)^{\mathrm{r}}$} & \makecell{$\underline{94.30}$\\$(0.44)^{\mathrm{r}}$} & \makecell{$\underline{98.67}$\\$(0.10)^{\mathrm{r}}$} & \makecell{$\underline{90.55}$\\$(0.45)^{\mathrm{r}}$} & \makecell{$88.53$\\$(1.51)^{\mathrm{s}}$} & \makecell{$65.93$\\$(1.28)^{\mathrm{b}}$} & \makecell{$\mathbf{57.47}$\\$(3.59)^{\mathrm{s}}$}\\
MASRouter & \makecell{$\underline{93.71}$\\$(0.23)^{\mathrm{r}}$} & \makecell{$93.80$\\$(0.17)^{\mathrm{r}}$} & \makecell{$98.50$\\$(0.10)^{\mathrm{r}}$} & \makecell{$88.98$\\$(0.39)^{\mathrm{r}}$} & \makecell{$88.40$\\$(1.59)^{\mathrm{s}}$} & \makecell{$63.14$\\$(1.31)^{\mathrm{b}}$} & \makecell{$56.00$\\$(3.33)^{\mathrm{s}}$}\\
GoAgent & \makecell{$92.87$\\$(0.04)^{\mathrm{r}}$} & \makecell{$\underline{94.30}$\\$(0.31)^{\mathrm{r}}$} & \makecell{$98.50$\\$(0.00)^{\mathrm{r}}$} & \makecell{$89.76$\\$(0.79)^{\mathrm{r}}$} & \makecell{$88.53$\\$(0.99)^{\mathrm{s}}$} & \makecell{$\mathbf{74.08}$\\$(1.19)^{\mathrm{b}}$} & \makecell{$56.07$\\$(2.89)^{\mathrm{s}}$}\\
VeriMap & \makecell{$93.63$\\$(0.39)^{\mathrm{r}}$} & \makecell{$93.60$\\$(0.25)^{\mathrm{r}}$} & \makecell{$\mathbf{98.83}$\\$(0.10)^{\mathrm{r}}$} & \makecell{$89.76$\\$(0.68)^{\mathrm{r}}$} & \makecell{$\underline{88.73}$\\$(0.50)^{\mathrm{s}}$} & \makecell{$65.71$\\$(0.28)^{\mathrm{r}}$} & \makecell{$56.33$\\$(3.87)^{\mathrm{s}}$}\\
ARGDes & \makecell{$93.63$\\$(0.38)^{\mathrm{r}}$} & \makecell{$\mathbf{94.60}$\\$(0.71)^{\mathrm{b}}$} & \makecell{$\underline{98.67}$\\$(0.17)^{\mathrm{r}}$} & \makecell{$\underline{90.55}$\\$(0.99)^{\mathrm{r}}$} & \makecell{$88.53$\\$(1.17)^{\mathrm{s}}$} & \makecell{$67.11$\\$(1.27)^{\mathrm{b}}$} & \makecell{$56.53$\\$(3.01)^{\mathrm{s}}$}\\
EIB & \makecell{$93.56$\\$(0.08)^{\mathrm{r}}$} & \makecell{$\underline{94.30}$\\$(0.10)^{\mathrm{r}}$} & \makecell{$98.50$\\$(0.10)^{\mathrm{r}}$} & \makecell{$89.76$\\$(0.39)^{\mathrm{r}}$} & \makecell{$\mathbf{88.80}$\\$(1.40)^{\mathrm{s}}$} & \makecell{$67.84$\\$(1.27)^{\mathrm{b}}$} & \makecell{$56.60$\\$(3.29)^{\mathrm{s}}$}\\
\bottomrule
\end{tabular}
\caption{Mathematical reasoning results. Each entry places the central score above its uncertainty term and source code. Source codes are defined above.}
\label{tab:full-main-results}
\label{tab:full-results-mathematical}
\end{table}

\begin{table}[H]
\centering
\footnotesize
\setlength{\tabcolsep}{2.0pt}
\renewcommand{\arraystretch}{1.10}
\begin{tabular}{lccccccc}
\toprule
Method & \makecell{MMLU\\Acc.\\$N=500$} & \makecell{MMLU-\\Redux\\Acc.\\$N=500$} & \makecell{MMLU-Pro\\Acc.\\$N=500$} & \makecell{StrategyQA\\Acc.\\$N=500$} & \makecell{Commonsense\\QA\\Acc.\\$N=500$} & \makecell{ARC-Easy\\Acc.\\$N=500$} & \makecell{ARC-\\Challenge\\Acc.\\$N=1418$}\\
\midrule
DeepSeek & \makecell{$85.40$\\$(0.92)^{\mathrm{s}}$} & \makecell{$83.73$\\$(0.50)^{\mathrm{s}}$} & \makecell{$73.07$\\$(0.76)^{\mathrm{s}}$} & \makecell{$84.60$\\$(0.80)^{\mathrm{s}}$} & \makecell{$85.40$\\$(0.53)^{\mathrm{s}}$} & \makecell{$\underline{94.27}$\\$(0.61)^{\mathrm{s}}$} & \makecell{$96.26$\\$(0.50)^{\mathrm{b}}$}\\
Qwen & \makecell{$82.33$\\$(2.76)^{\mathrm{s}}$} & \makecell{$79.80$\\$(0.92)^{\mathrm{s}}$} & \makecell{$60.93$\\$(0.90)^{\mathrm{s}}$} & \makecell{$79.40$\\$(1.93)^{\mathrm{s}}$} & \makecell{$85.67$\\$(1.33)^{\mathrm{s}}$} & \makecell{$94.07$\\$(0.76)^{\mathrm{s}}$} & \makecell{$96.05$\\$(0.52)^{\mathrm{b}}$}\\
\midrule
AutoGen & \makecell{$85.80$\\$(0.35)^{\mathrm{s}}$} & \makecell{$84.00$\\$(1.73)^{\mathrm{s}}$} & \makecell{$71.80$\\$(1.20)^{\mathrm{s}}$} & \makecell{$84.13$\\$(0.70)^{\mathrm{s}}$} & \makecell{$85.67$\\$(1.30)^{\mathrm{s}}$} & \makecell{$\underline{94.27}$\\$(0.76)^{\mathrm{s}}$} & \makecell{$96.61$\\$(0.15)^{\mathrm{r}}$}\\
MADebate & \makecell{$85.93$\\$(0.12)^{\mathrm{s}}$} & \makecell{$83.60$\\$(0.53)^{\mathrm{s}}$} & \makecell{$71.87$\\$(1.14)^{\mathrm{s}}$} & \makecell{$84.07$\\$(1.96)^{\mathrm{s}}$} & \makecell{$85.20$\\$(1.56)^{\mathrm{s}}$} & \makecell{$94.07$\\$(0.81)^{\mathrm{s}}$} & \makecell{$96.54$\\$(0.49)^{\mathrm{b}}$}\\
DyLAN & \makecell{$86.00$\\$(1.39)^{\mathrm{s}}$} & \makecell{$83.93$\\$(1.50)^{\mathrm{s}}$} & \makecell{$\mathbf{73.87}$\\$(1.21)^{\mathrm{s}}$} & \makecell{$\mathbf{86.67}$\\$(0.70)^{\mathrm{s}}$} & \makecell{$85.73$\\$(0.61)^{\mathrm{s}}$} & \makecell{$94.07$\\$(0.76)^{\mathrm{s}}$} & \makecell{$\underline{96.76}$\\$(0.19)^{\mathrm{r}}$}\\
\midrule
GPTSwarm & \makecell{$85.60$\\$(0.53)^{\mathrm{s}}$} & \makecell{$\underline{84.40}$\\$(0.35)^{\mathrm{s}}$} & \makecell{$70.73$\\$(0.61)^{\mathrm{s}}$} & \makecell{$85.67$\\$(2.84)^{\mathrm{s}}$} & \makecell{$85.53$\\$(0.64)^{\mathrm{s}}$} & \makecell{$\underline{94.27}$\\$(0.76)^{\mathrm{s}}$} & \makecell{$\mathbf{96.83}$\\$(0.19)^{\mathrm{r}}$}\\
GDesigner & \makecell{$85.67$\\$(1.51)^{\mathrm{s}}$} & \makecell{$84.27$\\$(1.17)^{\mathrm{s}}$} & \makecell{$71.87$\\$(0.31)^{\mathrm{s}}$} & \makecell{$85.00$\\$(2.03)^{\mathrm{s}}$} & \makecell{$86.13$\\$(1.36)^{\mathrm{s}}$} & \makecell{$94.20$\\$(1.06)^{\mathrm{s}}$} & \makecell{$96.33$\\$(0.50)^{\mathrm{b}}$}\\
SparseCT & \makecell{$85.20$\\$(0.87)^{\mathrm{s}}$} & \makecell{$83.73$\\$(0.50)^{\mathrm{s}}$} & \makecell{$71.33$\\$(1.10)^{\mathrm{s}}$} & \makecell{$85.80$\\$(0.80)^{\mathrm{s}}$} & \makecell{$86.13$\\$(1.21)^{\mathrm{s}}$} & \makecell{$93.93$\\$(0.64)^{\mathrm{s}}$} & \makecell{$96.69$\\$(0.48)^{\mathrm{b}}$}\\
GraphSearch & \makecell{$85.80$\\$(0.80)^{\mathrm{s}}$} & \makecell{$\underline{84.40}$\\$(1.91)^{\mathrm{s}}$} & \makecell{$71.60$\\$(0.92)^{\mathrm{s}}$} & \makecell{$85.60$\\$(0.20)^{\mathrm{s}}$} & \makecell{$86.07$\\$(1.27)^{\mathrm{s}}$} & \makecell{$\underline{94.27}$\\$(0.58)^{\mathrm{s}}$} & \makecell{$96.40$\\$(0.49)^{\mathrm{b}}$}\\
GraphR1 & \makecell{$85.33$\\$(1.30)^{\mathrm{s}}$} & \makecell{$84.00$\\$(1.06)^{\mathrm{s}}$} & \makecell{$70.53$\\$(1.33)^{\mathrm{s}}$} & \makecell{$85.60$\\$(0.69)^{\mathrm{s}}$} & \makecell{$85.47$\\$(1.67)^{\mathrm{s}}$} & \makecell{$94.20$\\$(0.72)^{\mathrm{s}}$} & \makecell{$\mathbf{96.83}$\\$(0.18)^{\mathrm{r}}$}\\
R-GFM & \makecell{$85.47$\\$(1.27)^{\mathrm{s}}$} & \makecell{$84.27$\\$(1.62)^{\mathrm{s}}$} & \makecell{$71.53$\\$(0.50)^{\mathrm{s}}$} & \makecell{$85.13$\\$(0.50)^{\mathrm{s}}$} & \makecell{$85.93$\\$(1.30)^{\mathrm{s}}$} & \makecell{$93.93$\\$(0.83)^{\mathrm{s}}$} & \makecell{$\underline{96.76}$\\$(0.36)^{\mathrm{r}}$}\\
BigMAS & \makecell{$86.00$\\$(1.11)^{\mathrm{s}}$} & \makecell{$\underline{84.40}$\\$(0.92)^{\mathrm{s}}$} & \makecell{$72.20$\\$(0.20)^{\mathrm{s}}$} & \makecell{$85.07$\\$(1.62)^{\mathrm{s}}$} & \makecell{$85.87$\\$(1.03)^{\mathrm{s}}$} & \makecell{$93.93$\\$(1.03)^{\mathrm{s}}$} & \makecell{$\mathbf{96.83}$\\$(0.16)^{\mathrm{r}}$}\\
MASRouter & \makecell{$\underline{86.53}$\\$(1.03)^{\mathrm{s}}$} & \makecell{$84.33$\\$(0.83)^{\mathrm{s}}$} & \makecell{$72.53$\\$(1.81)^{\mathrm{s}}$} & \makecell{$85.67$\\$(0.58)^{\mathrm{s}}$} & \makecell{$86.07$\\$(0.70)^{\mathrm{s}}$} & \makecell{$\mathbf{94.33}$\\$(0.64)^{\mathrm{s}}$} & \makecell{$96.61$\\$(0.25)^{\mathrm{r}}$}\\
GoAgent & \makecell{$85.93$\\$(1.14)^{\mathrm{s}}$} & \makecell{$83.87$\\$(1.33)^{\mathrm{s}}$} & \makecell{$70.33$\\$(0.50)^{\mathrm{s}}$} & \makecell{$85.27$\\$(2.05)^{\mathrm{s}}$} & \makecell{$\mathbf{86.47}$\\$(1.45)^{\mathrm{s}}$} & \makecell{$94.00$\\$(1.06)^{\mathrm{s}}$} & \makecell{$\underline{96.76}$\\$(0.18)^{\mathrm{r}}$}\\
VeriMap & \makecell{$85.87$\\$(0.46)^{\mathrm{s}}$} & \makecell{$84.13$\\$(1.51)^{\mathrm{s}}$} & \makecell{$\underline{73.53}$\\$(1.70)^{\mathrm{s}}$} & \makecell{$83.53$\\$(1.17)^{\mathrm{s}}$} & \makecell{$85.40$\\$(1.00)^{\mathrm{s}}$} & \makecell{$93.87$\\$(0.92)^{\mathrm{s}}$} & \makecell{$\underline{96.76}$\\$(0.12)^{\mathrm{r}}$}\\
ARGDes & \makecell{$\mathbf{87.27}$\\$(1.10)^{\mathrm{s}}$} & \makecell{$83.60$\\$(1.04)^{\mathrm{s}}$} & \makecell{$71.47$\\$(0.58)^{\mathrm{s}}$} & \makecell{$\underline{86.00}$\\$(1.59)^{\mathrm{s}}$} & \makecell{$\underline{86.20}$\\$(2.31)^{\mathrm{s}}$} & \makecell{$94.13$\\$(0.64)^{\mathrm{s}}$} & \makecell{$\underline{96.76}$\\$(0.25)^{\mathrm{r}}$}\\
EIB & \makecell{$86.13$\\$(1.67)^{\mathrm{s}}$} & \makecell{$\mathbf{84.53}$\\$(1.42)^{\mathrm{s}}$} & \makecell{$72.47$\\$(0.12)^{\mathrm{s}}$} & \makecell{$85.40$\\$(1.40)^{\mathrm{s}}$} & \makecell{$85.73$\\$(1.17)^{\mathrm{s}}$} & \makecell{$93.93$\\$(0.70)^{\mathrm{s}}$} & \makecell{$96.61$\\$(0.22)^{\mathrm{r}}$}\\
\bottomrule
\end{tabular}
\caption{Knowledge and commonsense results. Each entry places the central score above its uncertainty term and source code. Source codes are defined above.}
\label{tab:full-results-knowledge}
\end{table}

\begin{table}[H]
\centering
\footnotesize
\setlength{\tabcolsep}{4.0pt}
\renewcommand{\arraystretch}{1.10}
\begin{tabular}{lcccc}
\toprule
Method & \makecell{HumanEval\\Acc.\\$N=164$} & \makecell{HumanEval++\\Acc.\\$N=164$} & \makecell{LiveCodeBench\\v6\\Pass@$1$\\$N=175$} & \makecell{MultiAgentBench\\Coding\\CodeQ\\$N=102$}\\
\midrule
DeepSeek & $84.76\,(0.00)^{\mathrm{r}}$ & $\mathbf{80.49}\,(2.14)^{\mathrm{r}}$ & $20.57\,(2.01)^{\mathrm{r}}$ & $41.55\,(1.06)^{\mathrm{m}}$\\
Qwen & $\mathbf{90.24}\,(1.35)^{\mathrm{e}}$ & $76.83\,(0.70)^{\mathrm{r}}$ & $\mathbf{26.86}\,(1.39)^{\mathrm{e}}$ & $\mathbf{60.54}\,(4.84)^{\mathrm{b}}$\\
\midrule
AutoGen & $81.71\,(1.21)^{\mathrm{e}}$ & $73.78\,(3.43)^{\mathrm{b}}$ & $18.86\,(1.25)^{\mathrm{e}}$ & $42.01\,(4.89)^{\mathrm{b}}$\\
MADebate & $83.54\,(2.11)^{\mathrm{r}}$ & $77.44\,(2.46)^{\mathrm{r}}$ & $21.71\,(2.88)^{\mathrm{r}}$ & $40.69\,(4.86)^{\mathrm{b}}$\\
DyLAN & $85.98\,(1.19)^{\mathrm{e}}$ & $78.66\,(3.20)^{\mathrm{b}}$ & $22.86\,(1.23)^{\mathrm{e}}$ & $37.06\,(4.78)^{\mathrm{b}}$\\
\midrule
GPTSwarm & $84.76\,(0.93)^{\mathrm{r}}$ & $\mathbf{80.49}\,(5.21)^{\mathrm{r}}$ & $19.43\,(0.00)^{\mathrm{r}}$ & $41.91\,(4.89)^{\mathrm{b}}$\\
GDesigner & $85.37\,(1.21)^{\mathrm{e}}$ & $76.83\,(1.76)^{\mathrm{r}}$ & $\underline{23.43}\,(1.25)^{\mathrm{e}}$ & $39.46\,(4.84)^{\mathrm{b}}$\\
SparseCT & $84.76\,(1.83)^{\mathrm{r}}$ & $76.83\,(1.22)^{\mathrm{r}}$ & $21.14\,(1.44)^{\mathrm{r}}$ & $\underline{43.82}\,(4.91)^{\mathrm{b}}$\\
GraphSearch & $81.10\,(3.06)^{\mathrm{b}}$ & $74.39\,(1.06)^{\mathrm{r}}$ & $22.29\,(3.15)^{\mathrm{b}}$ & $40.64\,(4.86)^{\mathrm{b}}$\\
GraphR1 & $86.59\,(2.66)^{\mathrm{b}}$ & $75.61\,(1.06)^{\mathrm{r}}$ & $20.00\,(3.02)^{\mathrm{b}}$ & $38.09\,(4.81)^{\mathrm{b}}$\\
R-GFM & $89.02\,(1.16)^{\mathrm{e}}$ & $75.61\,(2.54)^{\mathrm{r}}$ & $19.43\,(0.87)^{\mathrm{r}}$ & $40.74\,(4.87)^{\mathrm{b}}$\\
BigMAS & $84.76\,(0.35)^{\mathrm{r}}$ & $76.22\,(3.32)^{\mathrm{b}}$ & $21.71\,(1.44)^{\mathrm{r}}$ & $37.79\,(1.04)^{\mathrm{m}}$\\
MASRouter & $76.83\,(3.29)^{\mathrm{b}}$ & $71.95\,(3.51)^{\mathrm{b}}$ & $20.57\,(0.57)^{\mathrm{r}}$ & $37.45\,(2.48)^{\mathrm{m}}$\\
GoAgent & $79.88\,(1.15)^{\mathrm{e}}$ & $74.39\,(2.46)^{\mathrm{r}}$ & $20.57\,(1.51)^{\mathrm{r}}$ & $37.79\,(4.80)^{\mathrm{b}}$\\
VeriMap & $78.66\,(3.20)^{\mathrm{b}}$ & $77.44\,(3.66)^{\mathrm{r}}$ & $21.14\,(0.33)^{\mathrm{r}}$ & $40.15\,(4.85)^{\mathrm{b}}$\\
ARGDes & $\underline{89.63}\,(1.18)^{\mathrm{e}}$ & $\underline{79.27}\,(3.17)^{\mathrm{b}}$ & $19.43\,(1.84)^{\mathrm{r}}$ & $39.90\,(4.85)^{\mathrm{b}}$\\
EIB & $83.54\,(2.20)^{\mathrm{r}}$ & $76.83\,(1.27)^{\mathrm{r}}$ & $21.71\,(1.44)^{\mathrm{r}}$ & $42.35\,(4.89)^{\mathrm{b}}$\\
\bottomrule
\end{tabular}
\caption{Code generation results. Entries give score (uncertainty)$^{\mathrm{source}}$ with the defined source codes. The recorded CodeQ metric is retained for MultiAgentBench-Coding.}
\label{tab:full-results-code}
\end{table}

\begin{table}[H]
\centering
\footnotesize
\setlength{\tabcolsep}{4.0pt}
\renewcommand{\arraystretch}{1.10}
\begin{tabular}{lcccc}
\toprule
Method & \makecell{PubMedQA\\Acc.\\$N=3000$} & \makecell{MedQA\\Acc.\\$N=1273$} & \makecell{MedMCQA\\Acc.\\$N=500$} & \makecell{MMLU-Med\\Acc.\\$N=1561$}\\
\midrule
DeepSeek & $81.33\,(0.71)^{\mathrm{b}}$ & $88.45\,(0.68)^{\mathrm{r}}$ & $79.67\,(0.99)^{\mathrm{s}}$ & $89.75\,(0.77)^{\mathrm{b}}$\\
Qwen & $80.70\,(1.25)^{\mathrm{b}}$ & $78.95\,(0.72)^{\mathrm{e}}$ & $73.53\,(0.61)^{\mathrm{s}}$ & $87.51\,(0.84)^{\mathrm{b}}$\\
\midrule
AutoGen & $81.43\,(0.71)^{\mathrm{b}}$ & $89.08\,(0.55)^{\mathrm{r}}$ & $79.67\,(1.70)^{\mathrm{s}}$ & $89.94\,(0.15)^{\mathrm{r}}$\\
MADebate & $81.87\,(0.70)^{\mathrm{b}}$ & $89.08\,(0.53)^{\mathrm{r}}$ & $78.53\,(1.14)^{\mathrm{s}}$ & $\underline{90.26}\,(0.75)^{\mathrm{b}}$\\
DyLAN & $\mathbf{82.43}\,(0.69)^{\mathrm{b}}$ & $88.77\,(0.08)^{\mathrm{r}}$ & $79.93\,(1.10)^{\mathrm{s}}$ & $89.56\,(0.38)^{\mathrm{r}}$\\
\midrule
GPTSwarm & $81.67\,(0.71)^{\mathrm{b}}$ & $89.63\,(0.54)^{\mathrm{e}}$ & $79.00\,(0.87)^{\mathrm{s}}$ & $89.94\,(0.24)^{\mathrm{r}}$\\
GDesigner & $81.37\,(0.71)^{\mathrm{b}}$ & $88.45\,(0.48)^{\mathrm{r}}$ & $79.93\,(0.76)^{\mathrm{s}}$ & $89.69\,(0.77)^{\mathrm{b}}$\\
SparseCT & $81.60\,(0.67)^{\mathrm{r}}$ & $89.24\,(0.21)^{\mathrm{r}}$ & $79.67\,(1.67)^{\mathrm{s}}$ & $89.88\,(0.39)^{\mathrm{r}}$\\
GraphSearch & $81.53\,(0.44)^{\mathrm{r}}$ & $\underline{89.71}\,(0.85)^{\mathrm{b}}$ & $78.73\,(1.01)^{\mathrm{s}}$ & $89.88\,(0.27)^{\mathrm{r}}$\\
GraphR1 & $81.30\,(0.71)^{\mathrm{b}}$ & $88.92\,(0.88)^{\mathrm{b}}$ & $79.13\,(0.83)^{\mathrm{s}}$ & $\mathbf{90.39}\,(0.75)^{\mathrm{b}}$\\
R-GFM & $81.53\,(0.21)^{\mathrm{r}}$ & $89.16\,(0.70)^{\mathrm{r}}$ & $79.53\,(0.95)^{\mathrm{s}}$ & $89.88\,(0.13)^{\mathrm{r}}$\\
BigMAS & $81.20\,(0.30)^{\mathrm{r}}$ & $\mathbf{89.87}\,(0.53)^{\mathrm{e}}$ & $78.73\,(1.85)^{\mathrm{s}}$ & $90.07\,(0.19)^{\mathrm{r}}$\\
MASRouter & $81.60\,(0.06)^{\mathrm{r}}$ & $89.24\,(0.33)^{\mathrm{r}}$ & $78.80\,(1.91)^{\mathrm{s}}$ & $89.56\,(0.10)^{\mathrm{r}}$\\
GoAgent & $80.53\,(0.72)^{\mathrm{b}}$ & $89.40\,(0.51)^{\mathrm{e}}$ & $79.13\,(0.92)^{\mathrm{s}}$ & $89.56\,(0.33)^{\mathrm{r}}$\\
VeriMap & $81.27\,(0.71)^{\mathrm{b}}$ & $88.77\,(0.52)^{\mathrm{r}}$ & $\underline{80.07}\,(1.51)^{\mathrm{s}}$ & $90.07\,(0.11)^{\mathrm{r}}$\\
ARGDes & $\underline{82.03}\,(0.70)^{\mathrm{b}}$ & $88.77\,(0.42)^{\mathrm{r}}$ & $\mathbf{80.53}\,(0.70)^{\mathrm{s}}$ & $89.62\,(0.77)^{\mathrm{b}}$\\
EIB & $81.20\,(0.46)^{\mathrm{r}}$ & $89.16\,(0.48)^{\mathrm{r}}$ & $78.87\,(1.33)^{\mathrm{s}}$ & $90.20\,(0.75)^{\mathrm{b}}$\\
\bottomrule
\end{tabular}
\caption{Biomedical reasoning results. Entries give score (uncertainty)$^{\mathrm{source}}$ with the defined source codes. PubMedQA retains the recorded source sample count.}
\label{tab:full-results-biomedical}
\end{table}

\begin{table}[H]
\centering
\footnotesize
\setlength{\tabcolsep}{4.0pt}
\renewcommand{\arraystretch}{1.10}
\begin{tabular}{lccc}
\toprule
Method & \makecell{TAT-QA\\Acc.\\$N=1663$} & \makecell{ConvFinQA\\Acc.\\$N=1490$} & \makecell{FinQA\\Acc.\\$N=1147$}\\
\midrule
DeepSeek & $70.48\,(0.63)^{\mathrm{r}}$ & $60.13\,(0.43)^{\mathrm{e}}$ & $38.10\,(0.51)^{\mathrm{e}}$\\
Qwen & $69.51\,(1.13)^{\mathrm{b}}$ & $\mathbf{61.81}\,(0.62)^{\mathrm{e}}$ & $28.51\,(0.70)^{\mathrm{e}}$\\
\midrule
AutoGen & $70.29\,(1.12)^{\mathrm{b}}$ & $60.74\,(0.66)^{\mathrm{r}}$ & $36.70\,(0.57)^{\mathrm{e}}$\\
MADebate & $70.66\,(1.12)^{\mathrm{b}}$ & $61.01\,(0.52)^{\mathrm{e}}$ & $36.97\,(0.61)^{\mathrm{e}}$\\
DyLAN & $70.48\,(0.23)^{\mathrm{r}}$ & $60.60\,(0.55)^{\mathrm{r}}$ & $\underline{38.88}\,(0.55)^{\mathrm{e}}$\\
\midrule
GPTSwarm & $70.41\,(0.33)^{\mathrm{r}}$ & $60.67\,(0.24)^{\mathrm{r}}$ & $37.66\,(1.13)^{\mathrm{r}}$\\
GDesigner & $70.23\,(0.44)^{\mathrm{r}}$ & $\underline{61.54}\,(0.48)^{\mathrm{e}}$ & $38.45\,(1.02)^{\mathrm{r}}$\\
SparseCT & $70.41\,(0.38)^{\mathrm{r}}$ & $60.81\,(0.47)^{\mathrm{e}}$ & $\mathbf{39.23}\,(0.56)^{\mathrm{e}}$\\
GraphSearch & $\mathbf{71.26}\,(1.11)^{\mathrm{b}}$ & $60.20\,(0.49)^{\mathrm{r}}$ & $38.45\,(0.18)^{\mathrm{r}}$\\
GraphR1 & $69.75\,(1.13)^{\mathrm{b}}$ & $60.67\,(0.34)^{\mathrm{r}}$ & $38.01\,(1.43)^{\mathrm{b}}$\\
R-GFM & $\underline{71.20}\,(1.11)^{\mathrm{b}}$ & $60.40\,(0.24)^{\mathrm{r}}$ & $38.45\,(0.46)^{\mathrm{r}}$\\
BigMAS & $70.41\,(0.68)^{\mathrm{r}}$ & $60.40\,(0.21)^{\mathrm{r}}$ & $38.36\,(0.52)^{\mathrm{e}}$\\
MASRouter & $69.99\,(0.24)^{\mathrm{r}}$ & $60.20\,(0.34)^{\mathrm{r}}$ & $37.14\,(1.43)^{\mathrm{b}}$\\
GoAgent & $71.14\,(1.11)^{\mathrm{b}}$ & $60.74\,(0.47)^{\mathrm{r}}$ & $38.45\,(0.13)^{\mathrm{r}}$\\
VeriMap & $70.23\,(0.33)^{\mathrm{r}}$ & $61.28\,(1.26)^{\mathrm{b}}$ & $37.66\,(0.83)^{\mathrm{r}}$\\
ARGDes & $71.02\,(1.11)^{\mathrm{b}}$ & $60.60\,(0.12)^{\mathrm{r}}$ & $38.45\,(0.31)^{\mathrm{r}}$\\
EIB & $69.99\,(0.67)^{\mathrm{r}}$ & $61.34\,(0.45)^{\mathrm{e}}$ & $38.54\,(0.54)^{\mathrm{e}}$\\
\bottomrule
\end{tabular}
\caption{Financial reasoning results with uncertainty terms and source codes.}
\label{tab:full-results-finance}
\end{table}

\begin{table}[H]
\centering
\footnotesize
\setlength{\tabcolsep}{4.0pt}
\renewcommand{\arraystretch}{1.10}
\begin{tabular}{lcccc}
\toprule
Method & \makecell{TabFact\\Acc.\\$N=500$} & \makecell{WikiSQL\\Acc.\\$N=500$} & \makecell{WikiTable\\Questions\\Acc.\\$N=500$} & \makecell{SQA\\Acc.\\$N=500$}\\
\midrule
DeepSeek & $92.00\,(1.11)^{\mathrm{s}}$ & $74.07\,(0.61)^{\mathrm{s}}$ & $76.73\,(2.16)^{\mathrm{s}}$ & $55.20\,(1.44)^{\mathrm{s}}$\\
Qwen & $81.40\,(1.56)^{\mathrm{s}}$ & $\underline{74.93}\,(1.27)^{\mathrm{s}}$ & $65.27\,(1.50)^{\mathrm{s}}$ & $46.53\,(2.14)^{\mathrm{s}}$\\
\midrule
AutoGen & $91.80\,(1.22)^{\mathrm{s}}$ & $74.20\,(1.78)^{\mathrm{s}}$ & $75.93\,(1.03)^{\mathrm{s}}$ & $54.67\,(1.97)^{\mathrm{s}}$\\
MADebate & $91.47\,(0.42)^{\mathrm{s}}$ & $74.27\,(1.10)^{\mathrm{s}}$ & $76.67\,(1.33)^{\mathrm{s}}$ & $54.60\,(1.93)^{\mathrm{s}}$\\
DyLAN & $\mathbf{92.53}\,(1.53)^{\mathrm{s}}$ & $73.87\,(2.53)^{\mathrm{s}}$ & $\underline{77.40}\,(1.44)^{\mathrm{s}}$ & $55.00\,(1.74)^{\mathrm{s}}$\\
\midrule
GPTSwarm & $92.20\,(1.40)^{\mathrm{s}}$ & $74.80\,(1.97)^{\mathrm{s}}$ & $76.60\,(1.93)^{\mathrm{s}}$ & $56.13\,(3.13)^{\mathrm{s}}$\\
GDesigner & $91.73\,(0.61)^{\mathrm{s}}$ & $\underline{74.93}\,(0.83)^{\mathrm{s}}$ & $76.60\,(1.06)^{\mathrm{s}}$ & $55.73\,(4.12)^{\mathrm{s}}$\\
SparseCT & $91.93\,(1.10)^{\mathrm{s}}$ & $74.87\,(1.22)^{\mathrm{s}}$ & $\mathbf{77.47}\,(0.76)^{\mathrm{s}}$ & $56.53\,(1.86)^{\mathrm{s}}$\\
GraphSearch & $92.00\,(1.56)^{\mathrm{s}}$ & $74.67\,(0.81)^{\mathrm{s}}$ & $77.13\,(1.62)^{\mathrm{s}}$ & $55.60\,(1.44)^{\mathrm{s}}$\\
GraphR1 & $91.40\,(1.71)^{\mathrm{s}}$ & $74.40\,(0.69)^{\mathrm{s}}$ & $76.80\,(0.20)^{\mathrm{s}}$ & $56.13\,(1.81)^{\mathrm{s}}$\\
R-GFM & $92.13\,(2.30)^{\mathrm{s}}$ & $\mathbf{75.60}\,(1.25)^{\mathrm{s}}$ & $\underline{77.40}\,(1.56)^{\mathrm{s}}$ & $56.07\,(3.67)^{\mathrm{s}}$\\
BigMAS & $\underline{92.33}\,(1.29)^{\mathrm{s}}$ & $74.73\,(0.90)^{\mathrm{s}}$ & $76.80\,(1.11)^{\mathrm{s}}$ & $55.33\,(3.03)^{\mathrm{s}}$\\
MASRouter & $91.40\,(1.11)^{\mathrm{s}}$ & $74.00\,(2.55)^{\mathrm{s}}$ & $75.67\,(0.81)^{\mathrm{s}}$ & $\underline{57.60}\,(1.51)^{\mathrm{s}}$\\
GoAgent & $\underline{92.33}\,(0.81)^{\mathrm{s}}$ & $74.47\,(1.79)^{\mathrm{s}}$ & $76.20\,(0.80)^{\mathrm{s}}$ & $\mathbf{57.73}\,(2.34)^{\mathrm{s}}$\\
VeriMap & $92.07\,(1.21)^{\mathrm{s}}$ & $74.33\,(1.63)^{\mathrm{s}}$ & $76.93\,(0.90)^{\mathrm{s}}$ & $54.87\,(1.27)^{\mathrm{s}}$\\
ARGDes & $92.07\,(0.83)^{\mathrm{s}}$ & $74.47\,(0.81)^{\mathrm{s}}$ & $77.00\,(1.60)^{\mathrm{s}}$ & $55.27\,(1.92)^{\mathrm{s}}$\\
EIB & $91.40\,(0.72)^{\mathrm{s}}$ & $74.60\,(1.59)^{\mathrm{s}}$ & $76.13\,(2.19)^{\mathrm{s}}$ & $55.53\,(3.16)^{\mathrm{s}}$\\
\bottomrule
\end{tabular}
\caption{Tabular reasoning results with uncertainty terms and source codes.}
\label{tab:full-results-tabular}
\end{table}

\begin{table}[H]
\centering
\footnotesize
\setlength{\tabcolsep}{4.0pt}
\renewcommand{\arraystretch}{1.10}
\begin{tabular}{lc}
\toprule
Method & \makecell{G-MAS-Complex\\Strict EM\\$N=400$}\\
\midrule
DeepSeek & $26.75\,(2.21)^{\mathrm{b}}$\\
Qwen & $13.25\,(1.70)^{\mathrm{b}}$\\
\midrule
AutoGen & $23.50\,(2.12)^{\mathrm{b}}$\\
MADebate & $14.00\,(1.73)^{\mathrm{b}}$\\
DyLAN & $4.50\,(1.04)^{\mathrm{b}}$\\
\midrule
GPTSwarm & $15.50\,(1.81)^{\mathrm{b}}$\\
GDesigner & $14.50\,(1.76)^{\mathrm{b}}$\\
SparseCT & $15.25\,(1.80)^{\mathrm{b}}$\\
GraphSearch & $14.00\,(1.73)^{\mathrm{b}}$\\
GraphR1 & $10.75\,(1.55)^{\mathrm{b}}$\\
R-GFM & $2.25\,(0.74)^{\mathrm{b}}$\\
BigMAS & $19.25\,(1.97)^{\mathrm{b}}$\\
MASRouter & $10.75\,(1.55)^{\mathrm{b}}$\\
GoAgent & $\mathbf{50.25}\,(2.50)^{\mathrm{b}}$\\
VeriMap & $\underline{29.50}\,(2.28)^{\mathrm{b}}$\\
ARGDes & $13.25\,(1.70)^{\mathrm{b}}$\\
EIB & $14.00\,(1.73)^{\mathrm{b}}$\\
\bottomrule
\end{tabular}
\caption{G-MAS-Complex results. Entries give strict exact match in percent with a binomial standard-error estimate in parentheses over 400 tasks.}
\label{tab:full-results-complex}
\end{table}

\FloatBarrier
\noindent\textbf{Sampling variability.} The following tables report means and sample standard deviations across three 500-instance sample sets. Their variation is across sampled task sets. These entries reproduce the cells marked $\mathrm{s}$ in the detailed results. Bold and underlined means identify the highest and second-highest distinct means in each dataset, including ties.

\begin{table}[H]
\centering
\footnotesize
\setlength{\tabcolsep}{4.0pt}
\renewcommand{\arraystretch}{1.10}
\begin{tabular}{lcc}
\toprule
Method & \makecell{ASDiv-A} & \makecell{MATH}\\
\midrule
DeepSeek & $88.53\!\pm\!1.10$ & $55.73\!\pm\!3.64$\\
Qwen & $86.87\!\pm\!0.42$ & $45.20\!\pm\!2.27$\\
\midrule
AutoGen & $88.53\!\pm\!0.99$ & $56.13\!\pm\!2.66$\\
MADebate & $88.53\!\pm\!0.99$ & $55.27\!\pm\!3.42$\\
DyLAN & $\mathbf{88.80}\!\pm\!1.40$ & $56.07\!\pm\!3.24$\\
\midrule
GPTSwarm & $88.67\!\pm\!0.81$ & $57.00\!\pm\!2.84$\\
GDesigner & $88.60\!\pm\!1.31$ & $56.73\!\pm\!3.37$\\
SparseCT & $\mathbf{88.80}\!\pm\!0.92$ & $\underline{57.13}\!\pm\!3.87$\\
GraphSearch & $88.67\!\pm\!0.90$ & $56.40\!\pm\!3.46$\\
GraphR1 & $88.60\!\pm\!1.06$ & $56.73\!\pm\!2.72$\\
R-GFM & $\mathbf{88.80}\!\pm\!1.44$ & $55.73\!\pm\!3.88$\\
BigMAS & $88.53\!\pm\!1.51$ & $\mathbf{57.47}\!\pm\!3.59$\\
MASRouter & $88.40\!\pm\!1.59$ & $56.00\!\pm\!3.33$\\
GoAgent & $88.53\!\pm\!0.99$ & $56.07\!\pm\!2.89$\\
VeriMap & $\underline{88.73}\!\pm\!0.50$ & $56.33\!\pm\!3.87$\\
ARGDes & $88.53\!\pm\!1.17$ & $56.53\!\pm\!3.01$\\
EIB & $\mathbf{88.80}\!\pm\!1.40$ & $56.60\!\pm\!3.29$\\
\bottomrule
\end{tabular}
\caption{Mathematical reasoning under sample-set variation. Entries report mean $\pm$ sample standard deviation in percent across three 500-instance sample sets.}
\label{tab:mean-std-results}
\label{tab:sampling-results-mathematical}
\end{table}

\begin{table}[H]
\centering
\footnotesize
\setlength{\tabcolsep}{2.0pt}
\renewcommand{\arraystretch}{1.10}
\begin{tabular}{lcccccc}
\toprule
Method & \makecell{MMLU} & \makecell{MMLU-\\Redux} & \makecell{MMLU-Pro} & \makecell{StrategyQA} & \makecell{Commonsense\\QA} & \makecell{ARC-Easy}\\
\midrule
DeepSeek & $85.40\!\pm\!0.92$ & $83.73\!\pm\!0.50$ & $73.07\!\pm\!0.76$ & $84.60\!\pm\!0.80$ & $85.40\!\pm\!0.53$ & $\underline{94.27}\!\pm\!0.61$\\
Qwen & $82.33\!\pm\!2.76$ & $79.80\!\pm\!0.92$ & $60.93\!\pm\!0.90$ & $79.40\!\pm\!1.93$ & $85.67\!\pm\!1.33$ & $94.07\!\pm\!0.76$\\
\midrule
AutoGen & $85.80\!\pm\!0.35$ & $84.00\!\pm\!1.73$ & $71.80\!\pm\!1.20$ & $84.13\!\pm\!0.70$ & $85.67\!\pm\!1.30$ & $\underline{94.27}\!\pm\!0.76$\\
MADebate & $85.93\!\pm\!0.12$ & $83.60\!\pm\!0.53$ & $71.87\!\pm\!1.14$ & $84.07\!\pm\!1.96$ & $85.20\!\pm\!1.56$ & $94.07\!\pm\!0.81$\\
DyLAN & $86.00\!\pm\!1.39$ & $83.93\!\pm\!1.50$ & $\mathbf{73.87}\!\pm\!1.21$ & $\mathbf{86.67}\!\pm\!0.70$ & $85.73\!\pm\!0.61$ & $94.07\!\pm\!0.76$\\
\midrule
GPTSwarm & $85.60\!\pm\!0.53$ & $\underline{84.40}\!\pm\!0.35$ & $70.73\!\pm\!0.61$ & $85.67\!\pm\!2.84$ & $85.53\!\pm\!0.64$ & $\underline{94.27}\!\pm\!0.76$\\
GDesigner & $85.67\!\pm\!1.51$ & $84.27\!\pm\!1.17$ & $71.87\!\pm\!0.31$ & $85.00\!\pm\!2.03$ & $86.13\!\pm\!1.36$ & $94.20\!\pm\!1.06$\\
SparseCT & $85.20\!\pm\!0.87$ & $83.73\!\pm\!0.50$ & $71.33\!\pm\!1.10$ & $85.80\!\pm\!0.80$ & $86.13\!\pm\!1.21$ & $93.93\!\pm\!0.64$\\
GraphSearch & $85.80\!\pm\!0.80$ & $\underline{84.40}\!\pm\!1.91$ & $71.60\!\pm\!0.92$ & $85.60\!\pm\!0.20$ & $86.07\!\pm\!1.27$ & $\underline{94.27}\!\pm\!0.58$\\
GraphR1 & $85.33\!\pm\!1.30$ & $84.00\!\pm\!1.06$ & $70.53\!\pm\!1.33$ & $85.60\!\pm\!0.69$ & $85.47\!\pm\!1.67$ & $94.20\!\pm\!0.72$\\
R-GFM & $85.47\!\pm\!1.27$ & $84.27\!\pm\!1.62$ & $71.53\!\pm\!0.50$ & $85.13\!\pm\!0.50$ & $85.93\!\pm\!1.30$ & $93.93\!\pm\!0.83$\\
BigMAS & $86.00\!\pm\!1.11$ & $\underline{84.40}\!\pm\!0.92$ & $72.20\!\pm\!0.20$ & $85.07\!\pm\!1.62$ & $85.87\!\pm\!1.03$ & $93.93\!\pm\!1.03$\\
MASRouter & $\underline{86.53}\!\pm\!1.03$ & $84.33\!\pm\!0.83$ & $72.53\!\pm\!1.81$ & $85.67\!\pm\!0.58$ & $86.07\!\pm\!0.70$ & $\mathbf{94.33}\!\pm\!0.64$\\
GoAgent & $85.93\!\pm\!1.14$ & $83.87\!\pm\!1.33$ & $70.33\!\pm\!0.50$ & $85.27\!\pm\!2.05$ & $\mathbf{86.47}\!\pm\!1.45$ & $94.00\!\pm\!1.06$\\
VeriMap & $85.87\!\pm\!0.46$ & $84.13\!\pm\!1.51$ & $\underline{73.53}\!\pm\!1.70$ & $83.53\!\pm\!1.17$ & $85.40\!\pm\!1.00$ & $93.87\!\pm\!0.92$\\
ARGDes & $\mathbf{87.27}\!\pm\!1.10$ & $83.60\!\pm\!1.04$ & $71.47\!\pm\!0.58$ & $\underline{86.00}\!\pm\!1.59$ & $\underline{86.20}\!\pm\!2.31$ & $94.13\!\pm\!0.64$\\
EIB & $86.13\!\pm\!1.67$ & $\mathbf{84.53}\!\pm\!1.42$ & $72.47\!\pm\!0.12$ & $85.40\!\pm\!1.40$ & $85.73\!\pm\!1.17$ & $93.93\!\pm\!0.70$\\
\bottomrule
\end{tabular}
\caption{Knowledge and commonsense under sample-set variation. Entries report mean $\pm$ sample standard deviation in percent across three 500-instance sample sets.}
\label{tab:sampling-results-knowledge}
\end{table}

\begin{table}[H]
\centering
\footnotesize
\setlength{\tabcolsep}{4.0pt}
\renewcommand{\arraystretch}{1.10}
\begin{tabular}{lc}
\toprule
Method & \makecell{MedMCQA}\\
\midrule
DeepSeek & $79.67\!\pm\!0.99$\\
Qwen & $73.53\!\pm\!0.61$\\
\midrule
AutoGen & $79.67\!\pm\!1.70$\\
MADebate & $78.53\!\pm\!1.14$\\
DyLAN & $79.93\!\pm\!1.10$\\
\midrule
GPTSwarm & $79.00\!\pm\!0.87$\\
GDesigner & $79.93\!\pm\!0.76$\\
SparseCT & $79.67\!\pm\!1.67$\\
GraphSearch & $78.73\!\pm\!1.01$\\
GraphR1 & $79.13\!\pm\!0.83$\\
R-GFM & $79.53\!\pm\!0.95$\\
BigMAS & $78.73\!\pm\!1.85$\\
MASRouter & $78.80\!\pm\!1.91$\\
GoAgent & $79.13\!\pm\!0.92$\\
VeriMap & $\underline{80.07}\!\pm\!1.51$\\
ARGDes & $\mathbf{80.53}\!\pm\!0.70$\\
EIB & $78.87\!\pm\!1.33$\\
\bottomrule
\end{tabular}
\caption{Biomedical reasoning under sample-set variation. Entries report mean $\pm$ sample standard deviation in percent across three 500-instance sample sets.}
\label{tab:sampling-results-biomedical}
\end{table}

\begin{table}[H]
\centering
\footnotesize
\setlength{\tabcolsep}{4.0pt}
\renewcommand{\arraystretch}{1.10}
\begin{tabular}{lcccc}
\toprule
Method & \makecell{TabFact} & \makecell{WikiSQL} & \makecell{WikiTable\\Questions} & \makecell{SQA}\\
\midrule
DeepSeek & $92.00\!\pm\!1.11$ & $74.07\!\pm\!0.61$ & $76.73\!\pm\!2.16$ & $55.20\!\pm\!1.44$\\
Qwen & $81.40\!\pm\!1.56$ & $\underline{74.93}\!\pm\!1.27$ & $65.27\!\pm\!1.50$ & $46.53\!\pm\!2.14$\\
\midrule
AutoGen & $91.80\!\pm\!1.22$ & $74.20\!\pm\!1.78$ & $75.93\!\pm\!1.03$ & $54.67\!\pm\!1.97$\\
MADebate & $91.47\!\pm\!0.42$ & $74.27\!\pm\!1.10$ & $76.67\!\pm\!1.33$ & $54.60\!\pm\!1.93$\\
DyLAN & $\mathbf{92.53}\!\pm\!1.53$ & $73.87\!\pm\!2.53$ & $\underline{77.40}\!\pm\!1.44$ & $55.00\!\pm\!1.74$\\
\midrule
GPTSwarm & $92.20\!\pm\!1.40$ & $74.80\!\pm\!1.97$ & $76.60\!\pm\!1.93$ & $56.13\!\pm\!3.13$\\
GDesigner & $91.73\!\pm\!0.61$ & $\underline{74.93}\!\pm\!0.83$ & $76.60\!\pm\!1.06$ & $55.73\!\pm\!4.12$\\
SparseCT & $91.93\!\pm\!1.10$ & $74.87\!\pm\!1.22$ & $\mathbf{77.47}\!\pm\!0.76$ & $56.53\!\pm\!1.86$\\
GraphSearch & $92.00\!\pm\!1.56$ & $74.67\!\pm\!0.81$ & $77.13\!\pm\!1.62$ & $55.60\!\pm\!1.44$\\
GraphR1 & $91.40\!\pm\!1.71$ & $74.40\!\pm\!0.69$ & $76.80\!\pm\!0.20$ & $56.13\!\pm\!1.81$\\
R-GFM & $92.13\!\pm\!2.30$ & $\mathbf{75.60}\!\pm\!1.25$ & $\underline{77.40}\!\pm\!1.56$ & $56.07\!\pm\!3.67$\\
BigMAS & $\underline{92.33}\!\pm\!1.29$ & $74.73\!\pm\!0.90$ & $76.80\!\pm\!1.11$ & $55.33\!\pm\!3.03$\\
MASRouter & $91.40\!\pm\!1.11$ & $74.00\!\pm\!2.55$ & $75.67\!\pm\!0.81$ & $\underline{57.60}\!\pm\!1.51$\\
GoAgent & $\underline{92.33}\!\pm\!0.81$ & $74.47\!\pm\!1.79$ & $76.20\!\pm\!0.80$ & $\mathbf{57.73}\!\pm\!2.34$\\
VeriMap & $92.07\!\pm\!1.21$ & $74.33\!\pm\!1.63$ & $76.93\!\pm\!0.90$ & $54.87\!\pm\!1.27$\\
ARGDes & $92.07\!\pm\!0.83$ & $74.47\!\pm\!0.81$ & $77.00\!\pm\!1.60$ & $55.27\!\pm\!1.92$\\
EIB & $91.40\!\pm\!0.72$ & $74.60\!\pm\!1.59$ & $76.13\!\pm\!2.19$ & $55.53\!\pm\!3.16$\\
\bottomrule
\end{tabular}
\caption{Tabular and structured reasoning under sample-set variation. Entries report mean $\pm$ sample standard deviation in percent across three 500-instance sample sets.}
\label{tab:sampling-results-tabular}
\end{table}

\FloatBarrier

\subsection{Artifact and Reproducibility}
\label{app:artifact-reproducibility}

The benchmark artifact separates task loading, baseline construction, backend execution, answer parsing, metric calculation, cost accounting, and intervention diagnosis. Each method is invoked through a common runner, making the raw prediction, parsed answer, metric result, trace, and cost fields comparable across single-agent, ordinary MAS, and graph-enhanced MAS configurations.

Each execution trace stores a method identifier, dataset identifier, sample identifier, organization hash, prompt-template hash, backend configuration, random seed when used, final prediction, parsed prediction, metric result, and cost fields. For every executed unit, the trace records role name, incoming message identifiers, outgoing message identifiers, model-call metadata, token counts, latency, retry status, and termination reason. For intervention runs, the trace additionally stores intervention parameters and changed graph or runtime elements.

\looseness=-1
The artifact specification includes dataset loaders, evaluation samples, baseline adapters, organization configurations, intervention operators, predictions, traces, cost logs, and aggregation scripts.

\textbf{Reproducibility records.}
\label{app:reproducibility-records-overview}
The artifact contains three linked record types.
A configuration record stores the dataset list, sample manifest, backend setting, prompt template, organization hash, and intervention parameters.
A scoring record stores the parser version, metric name, metric input, parsed prediction, invalid-parse flag, and score.
An aggregation record stores the included datasets, metric weights, uncertainty source, method grouping, and the rule for selecting representative configurations.
These records make the full tables auditable and preserve dataset-level metric identities even when results are summarized by domain or method family.

\section{Supplementary Organizational Diagnosis}
\label{app:organizational-diagnosis}

\subsection{Illustrative Multi-Document Task}
\label{app:complex-example}

This constructed example illustrates the dependency and conflict logic of G-MAS-Complex. The query asks for the available inventory of product P and the identifiers of the records supporting the calculation. The document collection contains the following records.

\begin{center}
\small
\begin{tabular}{@{}p{0.10\textwidth}p{0.81\textwidth}@{}}
\toprule
Record & Content \\
\midrule
D1 & Initial inventory of product P is 12 units. \\
D2 & Revised inventory of product P is 9 units. This record supersedes D1. \\
D3 & The reservation for product P is 4 units. \\
\bottomrule
\end{tabular}
\end{center}

The correct answer gives \texttt{available\_units} as 5. It sets \texttt{inventory\_source} to D2 and \texttt{reservation\_source} to D3. Extraction locates the records, revision checking selects D2, calculation obtains $9-4=5$, and verification checks the value and source identifiers. Two illustrative error cases distinguish these requirements. Selecting D1 produces $12-4=8$, while returning 5 without D3 omits a required source.

A collaboration graph can assign extraction, revision checking, calculation, and integration to separate units. The illustrative records expose the task's dependency structure and distinguish record selection, numerical calculation, and source reporting within the answer contract.

\subsection{Component Scoring and Input Contracts}
\label{app:component-scoring}

A diagnostic extension can decompose strict exact match into the following checks. Each component rate uses the full evaluated task list as its denominator. The full-task check retains all constraints of the original answer contract.

\begin{center}
\small
\begin{tabular}{@{}p{0.17\textwidth}p{0.74\textwidth}@{}}
\toprule
Component & Check \\
\midrule
Format & The output parses and contains the required fields and types. \\
Core answer & Requested values and rankings agree with the reference. \\
Source & Required source identifiers support the returned answer. \\
Conflict & Selected records follow the task's revision and authority rules. \\
Full task & All required checks, checkpoints, and checksum constraints hold. \\
\bottomrule
\end{tabular}
\end{center}

The input contract separates the query, source documents, dependency and conflict information, and answer schema from evaluator annotations. For $\Gamma_i$, it identifies relations stated in the input, relations encoded by document contents, and labels available to the evaluator. Each answer field has a type, a reference value, and a validation rule.

The construction record schema contains a task identifier, template identifier, generation procedure, and suite version. Generation fields specify the template or model revision, prompt, sampling settings, and random seed. Tier metadata describes document count, input length, dependency depth, conflicting records, distractors, and required answer fields. Validation fields specify reference and checksum checks, duplicate checks within and across splits, and human review records. Review fields identify the sampled tasks, annotators, decisions, and disagreement resolution. A frozen manifest joins accepted task identifiers with file hashes, the scoring revision, freeze date, and license.

\subsection{Reproducibility Record Schema}
\label{app:reproducibility-records}

The following schema defines the information needed to reconstruct a comparison. It specifies fields for dataset manifests, adapter configurations, and execution records. A comparison identifier joins these sources with predictions and scores.

\begin{center}
\small
\begin{tabular}{@{}p{0.19\textwidth}p{0.72\textwidth}@{}}
\toprule
Record group & Required fields \\
\midrule
Dataset & Version, split, filters, sample identifiers, order, sampling seed, frozen manifest, and file hashes. \\
Model & Exact version, request date, reasoning mode, decoding parameters, context limit, and output limit. \\
Adapter & Implementation revision, configuration hash, retained mechanisms, changes, role prompts, and stopping rules. \\
Learning & Training and validation manifests, test isolation, objective, checkpoint, and graph-generation procedure. \\
Tools & Tool permissions, verification access, candidate selection, and format-repair rules. \\
Shared state & State entries readable and writable by each unit, with message delivery and state access distinguished. \\
Run and score & Run identifier, random seed, parser and metric versions, score unit, aggregation rule, and reference answers. \\
Cost and failure & Tokens, calls, retries, latency, timeouts, parsing and tool failures, termination status, concurrency, caching, and hardware. \\
\bottomrule
\end{tabular}
\end{center}

For candidate replacement, the record includes eligible items, selected items, replacement content, and credibility markers in both reference and modified conditions. For worker failures, it includes the worker population, selection rule, injection times, duration, and whether the count represents simultaneous unavailability or accumulated events. A paired record joins reference and modified executions through the same task and configuration identifiers. It records the intervention type, selected items, random seed, schedule, and achieved modification level. Predictions, scores, and resource records share the execution identifier.

\subsection{Organization Analysis Protocol}
\label{app:organization-analysis}

Graph construction and execution describe different properties. GPTSwarm optimizes prompts and connectivity~\citep{zhuge2024gptswarm}. G-Designer learns task-dependent graph generation~\citep{zhang2024gdesigner}. ARG-Designer sequentially generates roles and links~\citep{li2026assemble}. Configuration records can describe these properties through the following attributes.

\begin{center}
\small
\begin{tabular}{@{}p{0.19\textwidth}p{0.72\textwidth}@{}}
\toprule
Attribute & Recorded configuration \\
\midrule
Topology & Fixed, generated for each task, or updated during execution. \\
Learning & Optimized prompts, graph parameters, routing rules, or policies. \\
Roles & Unit definitions and differences in prompts, tools, and models. \\
Verification & Checks available to each unit and their invocation rules. \\
Recovery & Retry, fallback, or rerouting rules and termination conditions. \\
\bottomrule
\end{tabular}
\end{center}

Graph records associate task identifiers with initial nodes, directed edges, updates, and executed messages. For a simple directed graph with $n\geq2$ nodes and $m$ edges, density is $m/[n(n-1)]$. Let $\mathcal{R}$ contain reachable ordered pairs of distinct nodes. The reachable-pair fraction and average reachable path length are
\begin{equation}
R_G=\frac{|\mathcal{R}|}{n(n-1)},
\qquad
L_G=\frac{1}{|\mathcal{R}|}\sum_{(u,v)\in\mathcal{R}}d_G(u,v).
\end{equation}
Here $d_G$ is shortest directed path length, and $L_G$ requires $|\mathcal{R}|>0$. Comparing constructed edges with executed message edges connects available routes with their use. For executions containing messages, let $p_v$ be the fraction originating from node $v$. Normalized source entropy is
\begin{equation}
H_G=-\frac{1}{\log n}\sum_{v\in V}p_v\log p_v,
\end{equation}
with zero-probability terms contributing zero. Role-level message counts and temporal records describe communication before and after an intervention.

\subsection{Reimplementation Validation Protocol}
\label{app:validation-protocol}

Reimplementation validation aligns the source revision, task split, backend, prompts, graph construction, optimization, decoding, stopping rule, and metric with the original method. Its comparison record contains the published score, reproduced score, their difference, and repetition count. A subsequent benchmark evaluation applies the shared task presentation and scoring controls. Method-specific roles and coordination rules remain part of the adapter description. This sequence connects original-setting validation with evaluation under the benchmark configuration.

\subsection{Follow-up Comparisons and Statistical Records}
\label{app:follow-up-protocols}

Future budget-matched comparisons can evaluate direct single-agent execution, iterative planning and verification, independent candidate generation, and representative multi-agent configurations under common model, tool, input, and resource limits. Actual consumption accompanies task scores. Candidate selection uses information available to the evaluated system. Role-replacement comparisons can preserve nodes and connections while replacing specialist prompts with general roles or exchanging role prompts. These settings test role definitions.

Further rewiring comparisons can check output reachability, scheduling validity, and schema compatibility for each proposed organization, alongside degree preservation. Repeated graph samples record seeds, accepted swaps, actual replacement ratios, and execution costs. Recovery comparisons can apply an identical failure schedule to the same configuration with recovery enabled and disabled, recording accuracy, retries, tokens, and latency.

Paired statistical analysis resamples common task identifiers and computes score differences within each replicate. The resampling specification groups generated tasks by template family to account for dependence among related instances. The analysis record specifies the resampling unit, repetition count, interval method, and central estimate. Separate run identifiers track execution randomness, while sample manifests track variation from dataset selection.

\FloatBarrier

\bibliographystyle{iclr2025_conference}
\setlength{\bibsep}{0pt}
\begingroup
\urlstyle{rm}
\let\originalbibitem\bibitem
\let\finishcompactentry\relax
\renewcommand{\bibitem}[2][]{%
  \finishcompactentry
  \originalbibitem[#1]{#2}%
  \begingroup
  \def\finishcompactentry{\par\endgroup}%
  \ifstrequal{#2}{chen-etal-2021-finqa}{\looseness=-1}{}%
  \ifstrequal{#2}{du2023multiagentdebate}{\looseness=-1}{}%
  \ifstrequal{#2}{zhu2025multiagentbench}{\looseness=-1}{}%
  \ifstrequal{#2}{jain2024livecodebenchholisticcontaminationfree}{%
    \lsstyle\spaceskip=2.3pt plus 1pt minus 0.7pt\looseness=-1}{}%
  \ifstrequal{#2}{wu2023autogen}{%
    \lsstyle\spaceskip=2.3pt plus 1pt minus 0.7pt\looseness=-1}{}%
  \ifstrequal{#2}{zhuge2024gptswarm}{%
    \lsstyle\spaceskip=2.3pt plus 1pt minus 0.7pt\looseness=-1}{}%
  \ignorespaces
}
\AtEndEnvironment{thebibliography}{\finishcompactentry}
\bibliography{reference}
\endgroup

\end{document}